\documentclass[journal]{IEEEtran}
\pdfoutput=1
\usepackage{times}

\usepackage[usenames, dvipsnames]{xcolor}
\usepackage[caption=false]{subfig}
\usepackage[numbers]{natbib}
\usepackage{multicol}
\usepackage[bookmarks=true, draft]{hyperref}
\usepackage[draft]{pgf}
\usepackage{amsmath}
\usepackage{comment}
\usepackage{booktabs}
\usepackage[inline]{enumitem}
\usepackage{tikz}
\usetikzlibrary{shapes,positioning,arrows}

\colorlet{mylinkcolor}{BrickRed}
\colorlet{mycitecolor}{Green}
\colorlet{myurlcolor}{NavyBlue}

\hypersetup{
  linkcolor  = mylinkcolor,
  citecolor  = mycitecolor,
  urlcolor   = myurlcolor,
  colorlinks = true,
}

\usepackage[markup=bfit, deletedmarkup=sout, authormarkup=brackets]{changes}

\graphicspath{{./gfx/}}

\usepackage[capitalize]{cleveref}

\newcommand\T{\rule{0pt}{2.6ex}}       %
\newcommand\B{\rule[-0.6ex]{0pt}{0pt}} %
\usepackage{array}
\newcolumntype{L}[1]{>{\raggedright\let\newline\\\arraybackslash\hspace{2pt}}m{#1}}
\newcolumntype{C}[1]{>{\centering\let\newline\\\arraybackslash\hspace{2pt}}m{#1}}
\newcolumntype{R}[1]{>{\raggedleft\let\newline\\\arraybackslash\hspace{2pt}}m{#1}}

\begin{document}

\title{How to be Helpful?\\ Implementing Supportive Behaviors\\ for Human-Robot Collaboration}

\author{Olivier~Mangin,~\IEEEmembership{Member,~IEEE,}
        Alessandro~Roncone,~\IEEEmembership{Member,~IEEE,}
        and~Brian~Scassellati,~\IEEEmembership{Member,~IEEE}%
\thanks{O. Mangin, A. Roncone and B. Scassellati are with the Social Robotics Lab,
        Computer Science Department, Yale University,
        51 Prospect St., New Haven, CT 06511, USA
        e-mail: name.surname@yale.edu}%
\thanks{Manuscript received Month Day, Year; revised Month Day, Year.}}

\maketitle

\begin{abstract}

The field of Human-Robot Collaboration (HRC) has seen a considerable amount of progress in the recent years. Although genuinely collaborative platforms are far from being deployed in real-world scenarios, advances in control and perception algorithms have progressively popularized robots in manufacturing settings, where they work side by side with human peers to achieve shared tasks. Unfortunately, little progress has been made toward the development of systems that are proactive in their collaboration, and autonomously take care of some of the chores that compose most of the collaboration tasks. In this work, we present a collaborative system capable of assisting the human partner with a variety of supportive behaviors in spite of its limited perceptual and manipulation capabilities and incomplete model of the task. Our framework leverages information from a high-level, hierarchical model of the task. The model, that is shared between the human and robot, enables transparent synchronization between the peers and understanding of each other's plan. More precisely, we derive a partially observable Markov model from the high-level task representation. We then use an online solver to compute a robot policy, that is robust to unexpected observations such as inaccuracies of perception, failures in object manipulations, as well as discovers hidden user preferences. We demonstrate that the system is capable of robustly providing support to the human in a furniture construction task.

\end{abstract}

\IEEEpeerreviewmaketitle

\section{Introduction} %
\label{sec:introduction}

\IEEEPARstart{R}{ecent} trends in advanced manufacturing and collaborative robotics are moving away from the traditional approach of building difficult-to-repurpose machines that work in isolation from humans. Rather, they are shifting focus to the development of mixed human-robot environments where robots are flexibly adaptable to the rapid changes of the modern manufacturing process and can safely and effectively inter-operate with humans. Unfortunately, albeit considerable progress in robot perception, manipulation and control has improved the robustness and dependability of such platforms, robots are still used as mere recipients of human instructions.
That is, we still lack the breadth and depth to fully exploit the capabilities of the robot in general and the collaboration in particular. Human-robot collaborative systems are fundamentally unbalanced, with the bulk of the perceptual, cognitive and manipulation capabilities still pertaining to the human side of the system.
For this reason, research in the field is shifting toward the deployment of systems that allow the human and the robot to focus on the tasks for which they are best suited, and mutually assist each other when needed~\citep{Shah2010,Breazeal2004,Hinds2004,Hayes2015}.

\begin{figure}
  \centering
  \includegraphics[width=.9\linewidth]{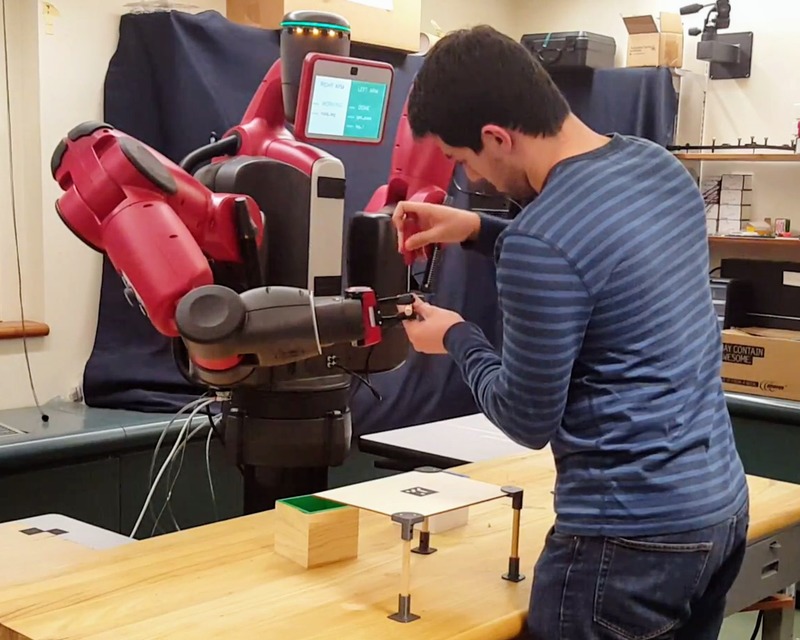}
  \caption{The experimental setup, in which a human participant engages in a joint construction task with the Baxter Robot. In the picture, the robot is supporting its human partner by holding a leg of the table while the human is screwing. See \cref{sub:experiment_design} for information about the task.}\label{fig:setup}
\end{figure}

In order for a robot to be proficient in the collaboration, we propose to focus our efforts on the implementation of platforms able to provide a variety of \textsl{supportive behaviors}. Our goal is for the robot to be capable of assisting the human \textsl{when the human needs support the most}, provided some degree of knowledge of the task at hand and sufficient information about the state of the system.
Supportive behaviors such as handing over task components, providing tools, cleaning-up unused elements, holding a part during assembly happen to be extremely beneficial for the completion of the task, and are within the realm of possibilities for modern robotic platforms. Such behaviors may also cover information retrieval tasks such as lighting an area of a working space, providing execution time, or reminding parts of the task-plan.
In such a scenario, it is not necessary for the robot to have exhaustive knowledge about the impending task, nor to be able to completely perceive the state of the world. Rather, for a system to exhibit effective support, partial observability about the state of the world and the internal state of the human partner (composed of her beliefs and intents) is sufficient.
Importantly, we do not set forth to improve the physical capabilities of our robotic platform, but rather we focus on how to optimally exploit the limited control, perceptive and reasoning skills the robot has in order to best support the human partner.

A key point for collaboration is that peers share a common understanding of the task, up to a certain degree~\cite{Searle1990,Shah2010}. Because of the gap between the current cognitive capabilities of humans and robots it is both troublesome for the robot to reach the level of task understanding that the humans have, and impractical for the humans---let alone non-technical persons---to encode that task model in a way the robot can exploit.
Therefore, we focus on basic models of the task, that are easy to describe for the human, yet contain enough information for the robot to be efficient and effective in its support. Notably, these shared task models can provide a substrate for human robot communication and thus foster transparent interactions between peers during task execution~\cite{Hoffman2004}.
Importantly, the robot is only made aware of the part of the task that matters to it, which simplifies both task planning and plan execution.
We do not aim to target robot controllers with exhaustive and exact knowledge---an approach that often falls short when outside of the context it has been designed for. Rather, we trade complete knowledge for adaptability, and optimal planning for good-enough support \textsl{by design}.

In this paper, we present a novel framework able to effectively empower a robot with supportive behaviors. We show, to our knowledge, the first practical implementation of an HRC application where the robot autonomously chooses to support the human when it deems it appropriate, and selects the right supportive action among the many it is provided with.
Our approach aims to maximize the throughput of the mixed human-robot system by leveraging the superior perceptual and manipulative skills of the human partner, while entrusting the robot with the role of autonomous helper, for which it is best suited for.
Additionally, we intend to systematically employ the human partner to retrieve the knowledge the robot is not able to gather by itself, in order to improve its estimation of the state of the world and the task completion.
Finally, our system is also able to adapt to user's preferences in terms of when and where to provide supportive behaviors.
In our experiments, we further demonstrate how the system is able to dynamically comply with users' preferences \textsl{during execution}.
We validate the proposed framework in joint construction tasks that aim to simulate the manufacturing process typical of Small and Medium Enterprises (SMEs), where features such as reconfigurability and ease of deployment are paramount.

In the following sections, we introduce the reader to the state of the art and related works in the field (\cref{sec:background_and_related_work}). Then, we detail the proposed approach, focusing on how it differentiates from relevant research in the topic (\cref{sec:method}). The experimental setup and the experiment design are presented in \cref{sec:implementation}, followed by the Results (\cref{sec:results}) and Conclusions (\cref{sec:discussion}).

\section{Background and Related Work} %
\label{sec:background_and_related_work}

In this work, we capitalize on past research in the field of high-level task reasoning and representation. As detailed in \cref{sec:method}, the core contribution of this paper is a system able to convert hierarchical task models into low-level planners capable of being executed by the robot.
Contrarily to more traditional techniques that leverage full observability in the context of HRC applications (e.g. \cite{Kaelbling2012,Toussaint2016}), our system deliberately optimizes its actions based on the interaction dynamics between the human and the robot. We explicitly account for uncertainty in the state of the world (e.g. task progression, availability of objects in the workspace) as well as in the state of the human partner (i.e. her beliefs, intents, and preferences). To this end, we employ a partially observable Markov decision process (POMDP) framework that plans optimal actions in the belief space.

To some extent, this approach builds on top of results in the field of task and motion planning (TAMP, see e.g. \cite{Kaelbling1998,Kaebling2013,Koval2016}).
Indeed, similarly to \citet{Kaebling2013} we find approximate solutions to large POMDP problems through planning in belief space combined with just-in-time re-planning.
Our work differs from traditional TAMP approaches in a number of ways:
\begin{enumerate*}[label=\roman*)]
  \item the hierarchical nature of the task is not explicitly dealt with in the POMDP model, but rather at a higher level of abstraction (that of the task representation, cf. \cref{sub:HTM}), which reduces complexity at planning stage;
  \item we %
  encapsulate the complexity relative to physically interacting with the environment away from the POMDP model, which results in broader applicability and ease of deployment if compared with standard TAMP methods;
  \item most notably, our domain of application presents fundamental differences with that targeted by TAMP techniques.
  That is, we propose to \textsl{handle uncertainty in the human-robot interaction}, rather than in the physical interaction between the robot and the environment.
  This latter point is worth elaborating on, since under this paradigm there is no shared consensus on how to model uncertainty about human's beliefs and intents in general, and the collaboration in particular.
\end{enumerate*}

Planning techniques can enable human robot collaboration when a precise model of the task is known, and might adapt to hidden user preferences as demonstrated by~\cite{Wilcox2012}. Similarly, partially observable models can provide robustness to unpredicted events and account for unobservable states.
Of particular note is the work by \citet{Gopalan2015} which, similarly to the approach presented in this paper, uses a POMDP to model a collaborative task.
Indeed, POMDPs and similar models (e.g. MOMDPs) have been shown to improve robot assistance~\cite{Hoey2010} and team efficiency~\cite{Nikolaidis2015} in related works.
Such models of the task are however generally expensive to build and require advanced technical knowledge.
Hence, a significant body of work in the fields of human-robot collaboration and physical human-robot interaction focuses on how to best take over the human partner by learning parts of the task that are burdensome in terms of physical safety or cognitive load.
Under this perspective, the majority of the research in the field has focused on frameworks for learning new skills from human demonstration (LfD, \cite{Billard2008}), efficiently learn or model task representations \cite{Hayes2016,Toussaint2016,Gombolay13,Ilghami2005}, or interpreting the human partner's actions and social signals~\cite{Grizou2013}.

No matter how efficient such models are at exhibiting the intended behavior, they are often limited to simple tasks and are not transparent to the human peer.
Indeed, evidences from the study of human-human interactions have demonstrated the importance of sharing mental task models to improve the efficiency of the collaboration~\cite{Shah2010}. Similarly, studies on human-robot interactions show that an autonomous robot with a model of the task shared with a human peer can decrease the idle time for the human during the collaboration~\cite{Shah2011}.
Without enabling the robot to learn the task, other approaches have demonstrated the essential capability for collaborative robots to dynamically adapt their plans with respect to the task in order to accommodate for human's actions or unforeseen events \cite{Hoffman2004}. Likewise, rich tasks models can also enable the optimization of the decision with respect to extrinsic metrics such as risk on the human~\cite{Hoffman2007} or completion time~\cite{Roncone2017}.

Our paper is positioned within this growing body of work related to task representations in HRC.
Unfortunately, little attention has been given to the issue of explicitly tackling the problem of effectively supporting the human partner.
To our knowledge, \citet{Hayes2015} is the only work that goes in this direction. It presents an algorithm to generate supportive behaviors during collaborative activity, although its results in simulation fall short in terms of providing practical demonstrations of the technique.
On the other side of the spectrum, a number of works cited above achieve to a certain amount supportive behaviors without explicitly targeting them \cite{Hoffman2007,Shah2011,Gopalan2015,Toussaint2016}.
A limitation of these approaches is that, as mentioned previously, they rely on exact task knowledge that is not always available for complex tasks in practical applications.

\section{Method}
\label{sec:method}

This work capitalizes on previous research by the authors. In \citet{Roncone2017}, we demonstrated an automated technique able to derive robot-executable policies from human-readable task models.
We then exploited this framework in the context of role assignment: our system was effective in efficiently negotiating allocation of a specific subtask to either the human or the robot during a collaborative assembly.
In this work, we expand this approach toward the more general problem of optimally providing support to the human.
Similarly to our previous work~\cite{Roncone2017}, we employ hierarchical representations of the task at a level of abstraction suitable to naive human participants and understandable by the robot. The model of the task is provided a priori, although other studies have shown that it is possible to learn task models from human demonstrations~\cite{Garland2003,Hayes2014}.
We then convert this task representation to a robot policy by leveraging the flexibility of POMDP models. This allows the robot to plan under uncertainty and explicitly reason at a high level of abstraction.

It is worth noting that exploiting adaptive planning from POMDPs has been already demonstrated in the context of human-robot collaboration (e.g.~\cite{Gopalan2015,Nikolaidis2015}).
Planning under uncertainty is indeed a major requirement for robots to interact with people. Near-future robotic platforms are most likely to operate in highly unstructured environments, for which even state of the art perception systems are not going to provide full observability or exact estimations.
In this work, we push this idea to its limits, inasmuch as we constrain our framework to the condition of being nearly \textsl{blind}.
That is, the robot is not able to directly observe neither the state of the world (task progression, object locations, etc.), nor the state of the partner (intents, preferences, etc).
This allows to investigate mechanisms of coordination though communication and physical interaction in the environment.
We achieve that by expanding the technique introduced in \cite{Roncone2017}, in that we leverage a high-level task model to \textsl{automatically generate} the lower-level POMDP. We then demonstrate how the resulting policy is successful in providing support during realistic human robot collaborations.

\subsection{Hierarchical Task Models (HTM)}
\label{sub:HTM}

Hierarchical structures form an appealing framework for high-level task representations; of particular interest is their capability to enable reuse of components over different tasks.
Additionally, their level of abstraction is usually close to human intuition: this facilitates human-robot communication about task execution~\cite{Roncone2017}.

\begin{figure*}
  \centering
  \subfloat[]{\includegraphics[width=.70\textwidth]{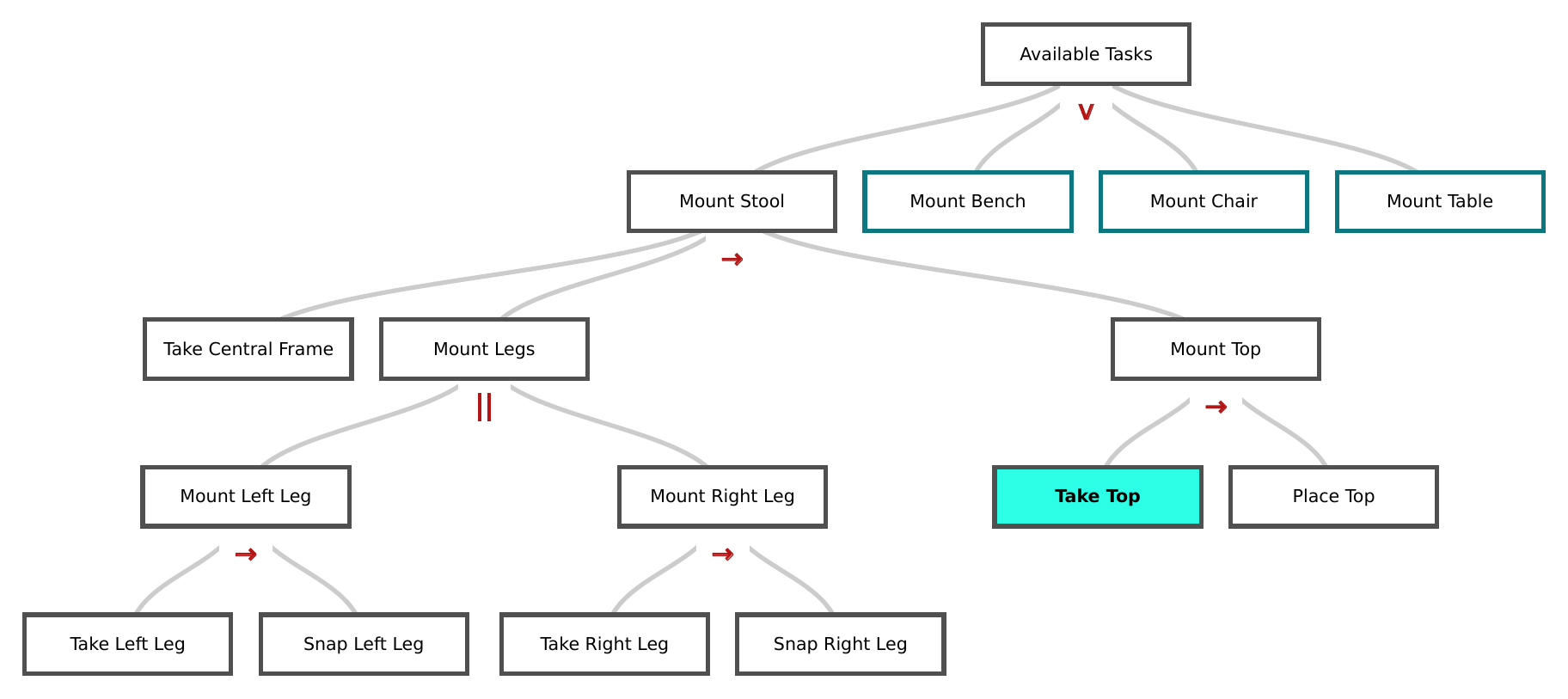}\label{fig:htm:icra}}\quad
  \subfloat[]{\includegraphics[width=.95\textwidth]{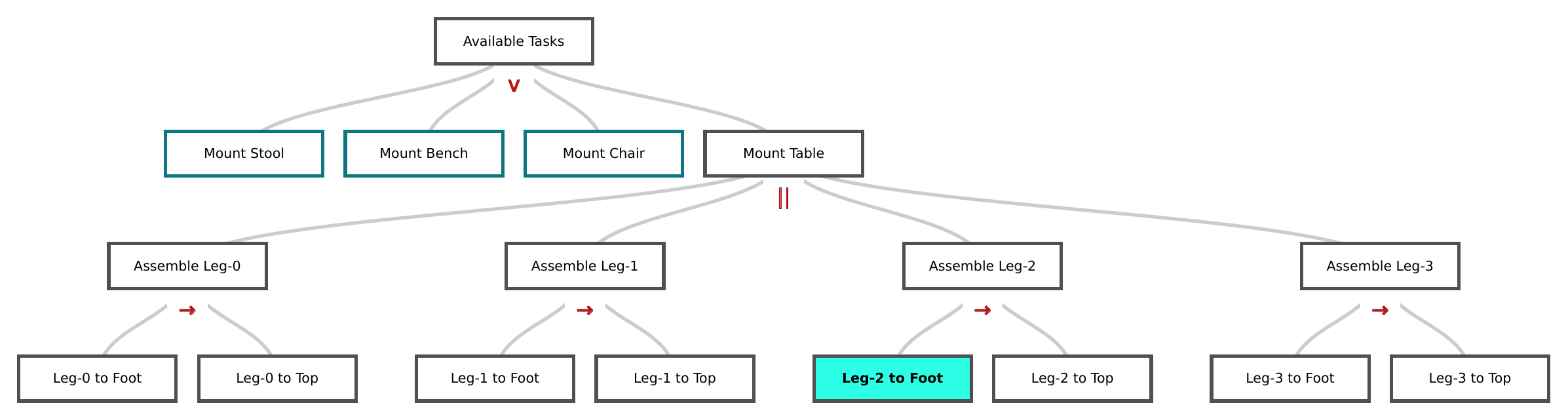}\label{fig:htm:supportive}}
  \caption{Example HTMs for human-robot assembly tasks. The user can retrieve information on the task execution, and inquire the system about task progression that will be highlighted by the robot during execution (cyan block in picture). The operator type between subtasks ($\rightarrow$ for sequential tasks, $||$ for parallel tasks, $\lor$ for alternative tasks) is also available for the user to inspect. \textbf{a)} HTM used in \citet{Roncone2017}. \textbf{b)} HTM used in \cref{sub:live_interaction}. See text for details.}
  \label{fig:htm}
\end{figure*}

\Cref{fig:htm} depicts some example representations for real-world construction tasks. Similarly to \citet{Hayes2016}, we consider HTMs built from primitive actions with operators that combine them into what we refer to as \textsl{subtasks} of increasing abstraction.
In this work, we assume that information about a set of primitive actions is already available to the robot, and we represent complex tasks on top of this action vocabulary.
We assume also that the robot has basic knowledge pertaining these primitive actions. This can range from knowing the type of tools and parts needed to perform an action, to being aware of the fact that supporting the human through holding a part may be beneficial during complex executions.
In our previous work~\cite{Roncone2017}, we extended the CC-HTM representation introduced in~\cite{Hayes2016} with the introduction of a new \textsl{alternative} operator ($\lor$ in \cref{fig:htm}). It adjoins the \textsl{sequential} ($\rightarrow$) and \textsl{parallel} ($||$) operators.
This set of operators proves suitable to capture the complexity of the collaboration, as well as the constraints of a task execution. For example, the parallel operator allows for the two peers to perform two disjoint subtasks as the same time; conversely, the sequential operator constrains them to a specific sequence of execution.

Thanks to their simplicity, HTM models can conveniently be drafted by non-expert workers and remain intuitive to understand.
Differently from traditional TAMP approaches~\cite{Kaelbling2012,Kaebling2013}, their high-level of abstraction also enables decoupling of the task planning component from the robot control element.
This increases flexibility, in that the robot just needs to be equipped with some motor primitives to match the atomic actions that compose the HTM.
From then on, the same library of motor primitives can be used to repurpose the robot to a new task.
Furthermore, it becomes easy to port the same task to a new platform with comparable motor and perception skills.

One of the main limitations of HTM-like approaches to HRC is that it is unlikely for the designer to be able to encode the totality of the information about the actual components of the task.
Although this poses limits to the breadth of applicability of such techniques, we argue that, for a robot to provide effective support, perfect knowledge about task execution is not needed in the first place.
A partial HTM, and specifically one that conveniently encodes only the information that matters to the robot, is sufficient for it to operate and interact with the human.
For example, the robot does not need to know how to perform a screwing action, nor how to perceive progression of the human screwing. What matters for a supportive robot is what objects are needed to complete said action, and that the human may be facilitated if the robot holds the part steadily.
As discussed in \cref{sub:POMDP}, our POMDP model complements partial knowledge about the task and the state of the world through interaction with the human partner.
It can for example supplant lack of perception about subtask progression by asking the human when the subtask is completed, or by directly moving on to the next subtask if its likelihood of subtask completion is high enough.

\subsection{Partially Observable Markov Decision Processes (POMDP)}
\label{sub:POMDP}

In this work, we use POMDPs to formulate the decision problem that the robot faces, given a task to solve collaboratively with a human and represented as an HTM as explained in \cref{sub:HTM}.
POMDPs are a generalization of Markov decision processes (MDPs), where there is only partial observability of the state of the process. This important relaxation of what defines an MDP allows for a significant gain in flexibility.
It is particularly relevant to model-imperfect perception and hidden states such as user preferences. We use such an approach to optimize the robot actions despite incomplete knowledge of the task and uncertainty regarding the dynamics of the collaboration.

More precisely, a POMDP is defined by a 7-tuple $(S,A,\Omega,T,O,R,\gamma)$, where $S$ is a set of states, $A$ is a set of actions, $T$ is a set of state transition probabilities, $R: S \times A \to {\rm I\!R}$ is the reward or cost function, $\Omega$ is a set of observations, $O$ is a distribution of observation probabilities, and $\gamma \in [0, 1]$ is the discount factor. Similarly to a MDP, at any given time the system lies in a specific state $s \in S$, which in the case of POMDPs is not directly observable.
The agent's action $a \in A$ triggers a state transition to state $s' \in S$ with probability $T(s'\mid s,a)$ and an observation $o \in \Omega$ with probability $O(o \mid s',a)$ that depends on the new state $s'$. Finally, the agent gets a reward $r \in \!R$ for taking the action $a$ while in state $s$.
In case of POMDPs, the agent's policy is defined on a probability distribution over states $b$, called the belief state, which accounts for the fact that the agent has no direct access to the real state $s$.
The goal is for the POMDP solver to find a policy $\pi(b): b \to a$ that maximizes the future discounted rewards over a possibly infinite horizon: $E \left[ \sum_{t=0}^\infty \gamma_t r_t \right]$.
Interestingly, actions that do not change the underlying state of the system but only the belief state are also valuable in this context.
This proves particularly beneficial for human collaboration, since information-gathering actions belong this paradigm.
For example, this intuition can be used to model communicative actions that trigger observations to disambiguate uncertainty, or to favor low-entropy beliefs with small uncertainty for both the human and the robot.
In reality, the belief state is usually very large and continuous; we use a policy that is defined on the history of previous actions and observations that we denote by $h\in H$. Please refer to \cref{sub:pomdp_planner} for more information on how we compute a robot policy from POMDP models.

\subsection{Restricted Model (RM)}
\label{ssub:restricted_model}

We propose an automated technique able to transform task-level HTMs into low-level robot policies through POMDPs.
To this end, we convert each primitive subtask (that is, each leaf composing the HTM in \cref{fig:htm:icra,fig:htm:supportive}) into a small, modular POMDP, which we call a restricted model (RM~\cite{Shani2014}). Hence, each RM is mostly independent from the rest of the problem and can be studied in isolation. Differently from ~\cite{Shani2014}, the RMs are composed at a later stage, according to the HTM structure, and the problem is solved in its entirety. This approach benefits from the modularity of the HTM representation, without the typical sub-optimality of policies that do not consider the full problem.

{%
\definecolor{yes}{RGB}{92,184,92}
\definecolor{no}{RGB}{56,148,240}
\definecolor{error}{RGB}{217,83,79}
\newcommand{\yes}{\textsl{yes} }
\newcommand{\no}{\textsl{no} }
\newcommand{\error}{\textsl{error} }
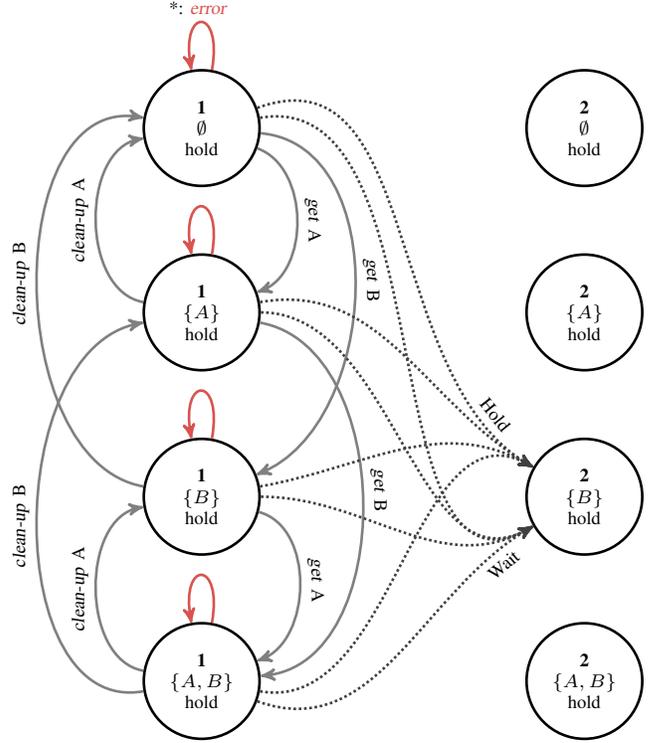
\begin{figure}
  \centering
    \begin{tikzpicture}[
      font=\scriptsize,
      >=stealth',
      ]

      \tikzstyle{state}=[
        draw=black,
        circle,
        text width=25pt,
        inner sep=4pt,
        text centered,
        font=\scriptsize,
        node distance=25pt and 100pt,
        line width=1,
      ]

      \tikzstyle{edge}=[line width=1, draw=gray]
      \tikzstyle{htm}=[draw=darkgray, densely dotted]
      \tikzstyle{label}=[sloped, inner sep=0pt, outer sep=3pt, auto]
      \tikzstyle{yes}=[draw=yes]
      \tikzstyle{no}=[draw=no]
      \tikzstyle{error}=[draw=error]

      \node[state] at (0,0)     (i)   {\textbf{1}\\$\emptyset$\\ hold};
      \node[state, below=of i]  (ia)  {\textbf{1}\\$\{A\}$\\ hold};
      \node[state, below=of ia] (ib)  {\textbf{1}\\$\{B\}$\\ hold};
      \node[state, below=of ib] (iab) {\textbf{1}\\$\{A,B\}$\\ hold};

      \node[state, right=of i]  (f)   {\textbf{2}\\$\emptyset$\\ hold};
      \node[state, below=of f]  (fa)  {\textbf{2}\\$\{A\}$\\ hold};
      \node[state, below=of fa] (fb)  {\textbf{2}\\$\{B\}$\\ hold};
      \node[state, below=of fb] (fab) {\textbf{2}\\$\{A,B\}$\\ hold};

      \draw[->, edge] (i)   to[in=20,  out=340] node[label, above] {\textsl{get} A}            (ia);
      \draw[->, edge] (ia)  to[in=190, out=170] node[label, above] {\textsl{clean-up} A}        (i);
      \draw[->, edge] (i)   to[in=22,  out=355] node[label, above, pos=0.48] {\textsl{get} B}  (ib);
      \draw[->, edge] (ib)  to[in=170, out=170] node[label, above] {\textsl{clean-up} B}        (i);
      \draw[->, edge] (ib)  to[in=20,  out=345] node[label, above, pos=0.48] {\textsl{get} A}           (iab);
      \draw[->, edge] (iab) to[in=190, out=170] node[label, above] {\textsl{clean-up} A}       (ib);
      \draw[->, edge] (ia)  to[in=5,   out=350] node[label, above, xshift=-5pt] {\textsl{get} B} (iab);
      \draw[->, edge] (iab) to[in=190, out=190] node[label, above] {\textsl{clean-up} B}       (ia);

      \draw[->, edge, error] (i)   to[in=100, out=80, loop] node[label, above] {*: {\color{error}\error}} (i);
      \draw[->, edge, error] (ia)  to[in=100, out=80, loop]                                              (ia);
      \draw[->, edge, error] (ib)  to[in=100, out=80, loop]                                              (ib);
      \draw[->, edge, error] (iab) to[in=100, out=80, loop]                                             (iab);

      \draw[->, edge, htm] (i)   to[in=150, out=20 ] node[label, very near end, above] {Hold} (fb);
      \draw[->, edge, htm] (i)   to[in=210, out=10 ]                                                  (fb);
      \draw[->, edge, htm] (ia)  to[in=150, out=10 ]                                                  (fb);
      \draw[->, edge, htm] (ia)  to[in=210, out=0  ]                                                  (fb);
      \draw[->, edge, htm] (ib)  to[in=150, out=10 ]                                                  (fb);
      \draw[->, edge, htm] (ib)  to[in=210, out=0  ]                                                  (fb);
      \draw[->, edge, htm] (iab) to[in=150, out=350]                                                  (fb);
      \draw[->, edge, htm] (iab) to[in=210, out=340] node[label, very near end, below] {Wait} (fb);

    \end{tikzpicture}
  \caption{Simplified representation of the restricted model (RM) used in this work. The figure represents an RM associated with a subtask \textbf{1} whose successor in the HTM is subtask \textbf{2}.
  For the sake of simplicity, the figure only represents actions that are taken starting from subtask \textbf{1}.
  We assume that only two objects `A' and `B' are available and that `A' is consumed by the subtask (like a part of the assembly would be) while `B' is a tool used during the task (e.g. a screwdriver).
  Each node represents a state, that is a factorization of the HTM subtask, each possible combination of objects on the workspace, and the user preference regarding the \textsl{hold} supportive action.
  Full connections in the graph represent successful transitions for the actions \textsl{get} and \textsl{clean-up} applied to objects `A' and `B' (that lead to a \textsl{none} observation).
  When taken from other states (e.g. bringing `A' which is already on the workspace), the action would fail, with an \textsl{error} observation, and the state would not change. These cases are represented by the red connections.
  Finally, the dotted connections represent the \textsl{hold} and \textsl{wait} actions that, from any of the represented states, lead to a transition to the state of the next subtask for which object `B' only is present (i.e. the tool).
  This means that the transition occurs even if the robot failed to bring all the required tools and parts: we assume here that the human would be able compensate for robot failures.
  The reward would however be maximal in the transition from state with $\{A, B\}$.
  To simplify the figure, we omitted the states corresponding to the \textsl{no-hold} preference.
  The graph for the \textsl{no-hold} preference is nearly identical except for the fact that the \textsl{hold} action fails from these states and hence \textsl{wait} is the only action to transition to the next subtask.}
  \label{fig:pomdp-subtask}
\end{figure}
}

\Cref{fig:pomdp-subtask} depicts the RM developed in this work.
As mentioned in \cref{sub:HTM}, its action space corresponds to the set of motor primitives available to the robot. For the purposes of this work, we consider the following supportive actions:
i) \textsl{wait} for the human to complete a subtask; ii) \textsl{hold} an object to provide support to the human; iii) \textsl{bring object} (e.g. constituent parts, small parts, buckets, or tools) onto the workspace; iv) \textsl{clean-up object} from the workspace when not needed anymore.
These motor primitives are implemented as independent controllers with their own logic: for example, the \textsl{wait} controller exploits communication in order to ask the human when the current operation has been completed before moving on to the new subtask. Modularity is employed in order to derive a distinct action for each object involved.

The set of possible observations is limited to a \textsl{none} observation (the default), plus a set of \textsl{error} observations returned by either the robot itself (e.g. \textsl{object-not-found}, \textsl{kinematic-error}) or the human partner (e.g. \textsl{wrong-action}).
Forcing the system to deal with a limited set of observation is intentional, and resorts to the intuition that, when the robot's execution is correct, the human partner should not be concerned about reinforcing this with positive feedback---which is time consuming and cognitively taxing.
Rather, feedback from the human partner should come either if explicitly requested by the robot---to disambiguate uncertainty, or if the robot's decision making is wrong---and negative feedback can be used to correct its course of actions.

Lastly, the state space $S$ is composed of a set of factored states. It is conceptually divided into three sub-spaces:
i) an \textsl{HTM-related state space} $\mathcal{Q}$ pertains to task progression, and is derived directly from the HTM representation. Each of the subtasks in the HTM (i.e. each of the leaves) is assigned an unique state $q \in \mathcal{Q}$;
additionally, one final state $\hat{q}$ is associated with a virtual operation that requires the robot to cleanup the workspace.
ii) a \textsl{subtask-related state space} is defined alongside the controllers that perform each subtask; it is composed of information relating parts and tools (e.g. if tools are present in the workspace, or if parts have been `consumed').
iii) a \textsl{human-related state space} encompasses information related to human preferences, beliefs and intents. As shown in \cref{sub:experimental_evaluation}, in this work we demonstrate how the proposed approach is able to adapt to user preferences regarding the \textsl{hold} supportive action.

It is worth noting how the size of the state space $S$ grows exponentially with the number of preferences and objects, as detailed in \cref{sub:pomdp_planner}.
In order to account for scalability of the method, we define a \textsl{generative POMDP model} that circumvents the issue of explicitly defining the full transition matrix $T$.
Instead, as detailed in \cref{fig:pomdp-subtask}, we generate it as follows. Each action affecting an object changes the state representing its presence in the workspace: for example, bringing a screwdriver makes it available on the workspace with high probability.
The \textsl{wait} action and eventually the \textsl{hold} action trigger the transition from one HTM leaf to the next (according to their order in the sequence, the final operations leading to a transition to the special state $\hat{q}$). This mechanism enforces that transitions between HTM states are transparently synchronized with the human. \textsl{Hold} only triggers the transition from states that have the preference for holding and fails otherwise.
Additionally, the transition from one leaf to the next erases from the state representation all the objects that have been `consumed' by the subtask (typically, the parts that have been used).
The initial state is sampled by starting at the initial subtask $q_0 \in \mathcal{Q}$; the workspace is assumed to be free of objects, and the human preferences are randomly set.

Interestingly, the design choice of limiting the perception of the state of the world naturally conforms to the statistical nature of a POMDP approach.
That is, adding uncertainty in the model makes it ultimately more robust to actual uncertainty in the collaborative interaction.
Without loss of generality, the RM can allow for unexpected transitions in order to account for actions of the human that are not observed by the robot---e.g. when the human partner fetches a required component by herself, unexpected failures, or missing objects.
To model this, and to avoid the robot to be stalled in a wrong belief, we introduce low probability random transitions between all the state features in $S$.

Finally, rewards are provided in the following cases: i) when the robot proposes to hold, and the human has preference for holding; ii) when there is a transition between a subtasks and its successor; ii) at completion of the full task. Additionally, each action taken by the robot has an intrinsic cost, and a negative reward is also given when the human has to bring or clean an object which was not taken care of by the robot. \Cref{tab:rewards} provides a summary of the rewards used in the experiment from \cref{sub:live_interaction}.

{%
\begin{table}
  \caption{Rewards used to train the policy on the POMDP model derived from the HTM.
  Instead of a table of rewards for all state transitions, actions, and observations, we present the rewards as triggered by events that can be cumulated.
  For example, if a final state is reached through a wait action, but the screwdriver is still on the workspace, a reward of $85$ is obtained.}
  \centering
  \begin{tabular}{lc}
    \toprule
    \textbf{Event} & \textbf{Reward} \T \B \\
    \midrule
    Final state reached                      & $100$ \\
    Subtask transition                       & $10$  \\
    Missing tool or part on state transition & $15$  \\
    Uncleaned object on final state          & $15$  \\
    User preference is honored               & $10$  \\
    Hold action taken                        & $-2$  \\
    Wait action taken                        & $0$   \\
    Other action taken                       & $-1$  \\
    \bottomrule
  \end{tabular}\label{tab:rewards}
\end{table}
}

\subsection{POMDP Planner}
\label{sub:pomdp_planner}

In this work, we implement a planner based on POMCP~\cite{Silver2010}, which is able to plan from generative models and can handle very large state spaces.
This is achieved through Monte-Carlo estimation of the beliefs by using a set of particles to represent each belief. A policy is learned based on all the visited histories, which constrains exploration to feasible states only.
Using particles for belief representation and Monte-Carlo techniques for value estimation addresses the issue of the belief space being too large to be explicitly represented in its entirety.
In a realistic HRC domain --such as those detailed in \cref{sec:results}-- there are typically thousands of states and tens of actions, but the amount of plausible states at any time is limited.
Hence, representing the $|\mathcal{S}|$-dimensional belief is not feasible, but at the same time, despite large state spaces, beliefs are sparse.
This is well represented through sets of particles, that naturally conform to sparse representations.
Further, this approach only requires a generative model of the transitions instead of representing the full transition matrix, whose dimension is $|\mathcal{S}|^2 \times |\mathcal{A}| \times |\mathcal{O}|$.

The computational complexity of the planning is bound to an exploration-exploitation trade-off. Namely, the planner explores a tree whose branching factor is equal to the number of actions multiplied by the number of observations.
Therefore, an important parameter to control the complexity of the problem is the horizon of the exploration.
More precisely, we define the horizon either in terms of the number of transitions or as the number of \textsl{HTM subtasks} that the exploration accounts for.
Such optimizations result in locally optimal decisions for what concerns a fixed number of subtasks.%

In order to limit computational complexity, it is also possible to remove part of the randomness introduced in \cref{ssub:restricted_model} for what concerns the transitions and the observations---although it still occurs in the belief transitions that hence still represent all the possible hypothesis.
By doing so, we prevent the planner from exploring feasible but rare events.
In case these rare events occur during live interaction with with the human, the online component of the planner is able to re-compute a new policy.
This makes the algorithm robust to unexpected events without penalizing the exploration.
That is, we artificially simplify the model of the interaction at the offline planning step, but we then compensate for its imperfections online---during task execution.
\section{Implementation} %
\label{sec:implementation}

\subsection{Experimental Setup} %
\label{sub:experimental_setup}

The experimental evaluation is carried out on a Baxter Research Robot (cf. \cref{fig:setup}), using the Robot Operating System (ROS \cite{ROS2009}). As mentioned in \cref{sec:method}, even though we do not concern with improving the physical capabilities of the platform, we leverage the state of the art in robot perception and control in order to build up a set of basic capabilities for the robot to effectively support its human partner.
The resulting framework, originally developed in \citet{Roncone2017}, exposes a library of high level actions, that are the only interface through which the POMDP planner can send information to---and retrieve information from---the Baxter system and the experimental setup.

The system presented in \cite{Roncone2017} provides multiple, redundant communication channels to interact with the human partner `\textsl{on human terms}'\cite{Breazeal2002}.
Among the list of available layers, for the purposes of this work we employ:
i) a \textsl{Text-to-Speech (TTS)} channel, used to verbally interact with the human;
ii) a \textsl{Feedback} channel, shown in the robot's head display, which provides feedback about its internal states and intents (see \cref{fig:setup});
iii) an \textsl{Error} channel that allows the human to send error messages to the robot, and is triggered by pressing one of the buttons on the robot's end effectors.
In addition to this, a fourth channel has been implemented, in the form of a \textsl{Speech-to-Text (STT)} system able to convert human sentences into robot-readable commands. It employs the Google Cloud STT API \cite{GoogleSTT} combined with a text parser that relies on a dictionary shared in advance with the human participant.

\begin{figure}
\centering
  \subfloat[Left  end effector's camera view, which is partly obscured by the vacuum   gripper (top part of picture).]{\includegraphics[width=.8\linewidth]{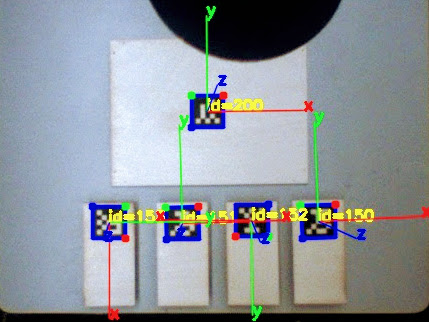} \label{fig:robot:left}}\\
  \subfloat[Right end effector's camera view, which is partly obscured by the parallel gripper (top part of picture).]{\includegraphics[width=.8\linewidth]{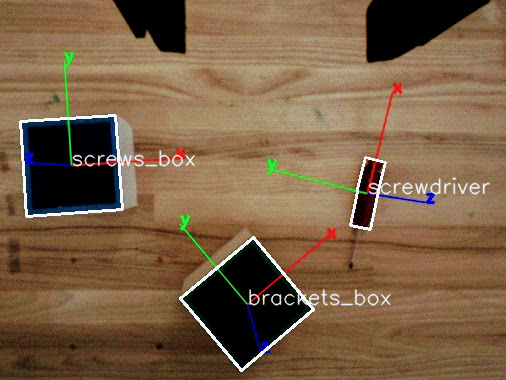} \label{fig:robot:right}}

  \caption{Snapshot of the camera streams from the left (\cref{fig:robot:left}) and right (\cref{fig:robot:right}) end effectors. The left arm uses a fiducial marker tracking system based on \cite{Aruco2014}, whereas the right arm implements the HSV-based 3D reconstruction software detailed in \cref{ssub:6d_object_reconstruction_from_single_view}. See \cref{sub:experiment_design} below for a description of the objects composing the construction task.}\label{fig:robot}
\end{figure}

Both arms are able to perform precise, closed loop visual servoing tasks thanks to a pair of cameras their end effectors are equipped with. The left arm is equipped with a vacuum gripper, able to pick up flat surfaces with constant texture, whereas the right end-effector is a parallel electric gripper capable of performing more complex grasping tasks.
We maximize the usage of both arms by leveraging their respective embodiments: to this end, two different perceptual systems have been employed (cf. \cref{fig:robot}). The perception system for the left arm (\cref{fig:robot:left}) is provided by ARuco \cite{Aruco2014}, a library capable of generating and detecting fiducial markers that are particularly suitable for being positioned on flat surfaces.
For what concerns the right arm, a custom color-based pose estimation algorithm has been implemented. It is detailed in the following section.

\subsubsection{6D object reconstruction from single view} %
\label{ssub:6d_object_reconstruction_from_single_view}

We consider here the scenario in which the end-effector is vertically placed on top of the pool of objects, and the objects are in the field of view of the camera. In order to be able to grasp a variety of objects with the parallel gripper installed on the Baxter's right arm (see \cref{fig:robot:right}), the following two steps are to be performed: objects need to be firstly detected in the camera view, and then their position and orientation has to be reconstructed in the 3D operational space of the robot.
For what concerns the former, a number of different computer vision techniques can be employed. In this work, we utilize a Hue-Saturation-Value (HSV) color segmentation algorithm: that is, each object is detected thanks to its color in the HSV color space, and its bounding box is stored for later use.
After an object has been detected, it is necessary to estimate its 3D pose in the world reference frame. We assume here that the object's physical sizes (width and height) are known, and that the matrices of intrinsic and extrinsic parameters $K$ and $\left[ R | T \right]$ are available.
Notably, whilst $K$ can be estimated via a prior camera calibration step, the extrinsic parameters are computed thanks to the robot's kinematics and the knowledge of the current joint configuration.
In this context, a standard perspective transformation can be applied in order to estimate the 3D position of a point $P_w = \left[ X\ Y\ Z\ 1\right]^T$ in the world reference frame from its corresponding image point $p_c = \left[ u\ v\ 1\right]^T$ in the camera reference frame. The following equation holds:

\begin{equation}
  s\ p_c = K \left[ R | T \right]\ P_w\ ,\label{eq:persp_trans_1}
\end{equation}

where $s$ is a scale factor\footnote{ It is out of the scope of this work to present the perspective transformation problem in detail. Please refer to \citet{Hartley2004} for more information on the topic.}.
The perspective transformation equation is then applied to estimate the pose of the mass center of the object by iteratively minimizing the reprojection error of its corners via a Levenberg-Marquardt algorithm \cite{Marquardt1963}, using the OpenCV computer vision library \cite{OpenCV2000}.

The technique proposed here is subject to a number of estimation and computational errors, in particular if the distance between the camera and the desired object is significant. In spite of that, we capitalize on the fact that the algorithm is employed in a visual servoing setup, in which the robot refines its estimation the closer it gets to the object.
That is, even if the initial pose reconstruction may be defective, it is continuously updated with a frequency of $30$ Hz and refined until the end effector reaches the object.
The authors acknowledge that more advanced 3D reconstruction techniques could be used, such as exploiting a depth sensing camera properly calibrated with respect to the robot---e.g. \cite{Jiang2011}.
The main advantage of the proposed solution is however to employ a compact, self-contained estimation step that does not rely on external equipment or burdensome calibration.
We consider this an important asset of our approach, that facilitates re-use and applicability to novel domains.

\subsection{Experiment Design} %
\label{sub:experiment_design}

\begin{figure}
\centering
  \subfloat[Parts composing the table.]{\includegraphics[width=0.8\linewidth]{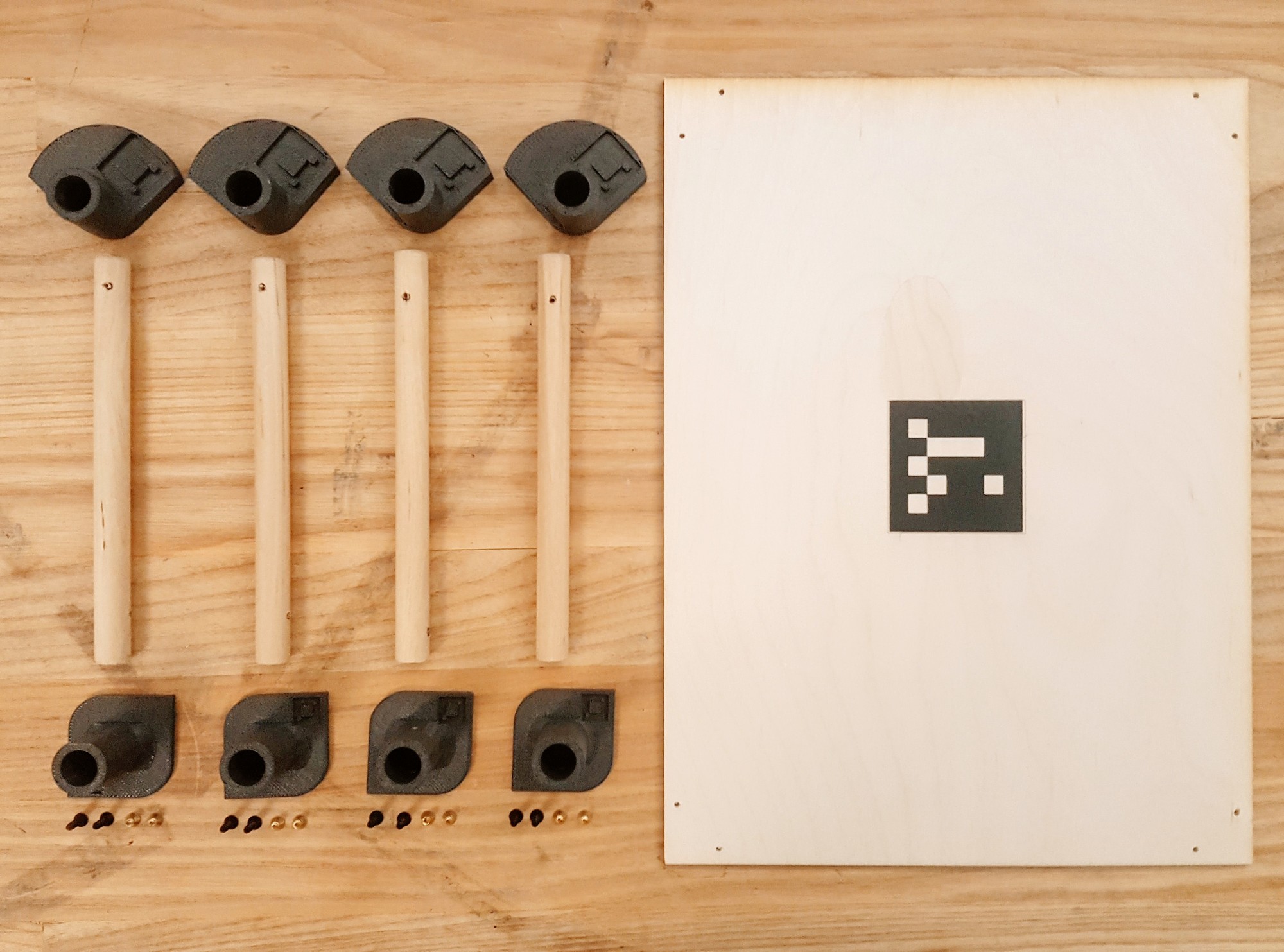} \label{fig:table:parts}}\\
  \subfloat[Completed table.]          {\includegraphics[width=0.8\linewidth]{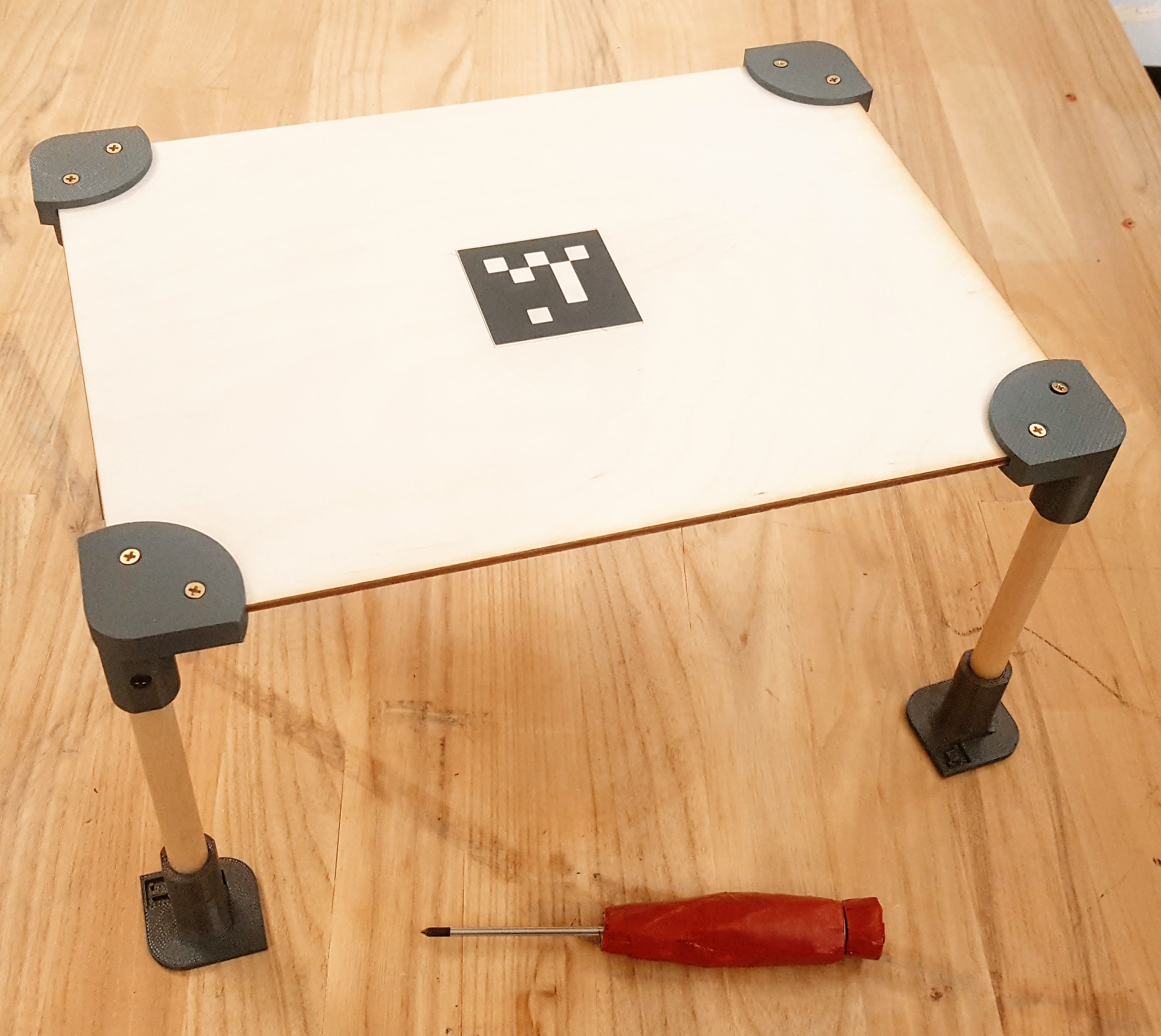} \label{fig:table:table}}

  \caption{\textbf{a)} The table building task is composed of one plywood tabletop, four dowels that act as legs, four brackets (top 3D printed objects in figure) and four feet (bottom 3D printed objects in figure). A total of 16 screws are needed to secure parts together. Both the tabletop and the legs have been pre-drilled to facilitate assembly. \textbf{b)} The table after the construction task is completed. The only tool required for the assembly is a screwdriver (red object in bottom of \cref{fig:table:table}).}\label{fig:table}
\end{figure}

As detailed in \cref{sec:introduction,sub:experimental_setup}, we perform our experiments in a collaborative scenario where human participants engage in a construction task with the Baxter Robot (see \cref{fig:setup}).
The collaborative task the two peers are engaged with is the joint construction of a miniaturized table (cf. \cref{fig:table}). It is composed of five structural elements---the tabletop and four legs---and eight custom 3D-printed linkages---four brackets secure the legs to the tabletop, whereas four feet are used to stabilize the structure.
A total of $16$ screws are required for the assembly. A screwdriver is the only tool needed to build the table, which has an approximate size of $30 \times 21 \times 15$ cm when completed.

Particular attention has been placed in the conceptualization of the task. The main goal was to tailor the design of the table to the typical constraints human-robot collaboration experiments present.
We purposely aim for: i) ease of reuse; ii) ease of retrieval of the constituent parts; iii) scalability; iv) proximity with real-world HRC applications, that are typically characterized by a combination of complex actions to be performed by the human partner (which often involve the use of tools), and simpler tasks the robot is usually assigned to.
The choices taken at the design stage allow us to comply with these requirements: the brackets used in this experiment belong to a larger library of linkages that has been made available online\footnote{ \href{https://scazlab.github.io/HRC-model-set}{\texttt{scazlab.github.io/HRC-model-set}} hosts CAD models, specifications for 3D printing, tutorials for assembling example designs, and reference links for purchasing parts.}, whereas the table shown in \cref{fig:table:table} is only one of the many designs allowed by our solution.
We plan on leveraging these features in future works, and to promote this design as a model set to benchmark human-robot collaboration.

For the purposes of this work, the two partners have distinct, non overlapping roles: the human (hereinafter referred to as the \textsl{builder}) is in charge of performing actions that require fine manipulation skills (e.g. screwing) or complex perception capabilities (such as inserting the top of the table onto a bracket); the robot (also called the \textsl{helper}) is instructed to back the builder with the supportive actions described below.
Importantly, the flexibility of the POMDP planner allows for a certain slack in terms of role assignment and task allocation by design. As detailed in \cref{sec:method}, the planner is automatically able to comply with unlikely states of the system, and to re-plan accordingly.
As a practical consequence to this, we are in the position of allowing the human participant to take charge of some supportive actions if he so chooses.
That is, we disclose to the builder that she is allowed to retrieve parts and tools for herself, even though we do not enforce that on the user---nor we convey that it is part of her duties as a participant of the experiment.

The robot helper is provided with a set of basic capabilities encapsulated into a library of high-level actions.
The supportive actions it has been instructed to perform are the following: i) retrieve parts (e.g. tabletop, screws, or legs); ii) retrieve the tool (namely, the screwdriver); iii) cleanup the workplace from the objects that are not going to be needed in the future; iv) hold structural parts in order to facilitate the builder's actions.
To comply with the Baxter's limited manipulation capabilities, we positioned the smaller components (i.e. the screws and the 3D printed objects) in apposite boxes, to be picked up by the parallel gripper---see \textsl{screws\_box} and \textsl{brackets\_box} in \cref{fig:robot:right}.
Similarly, the legs of the table have been equipped with a specific support in order to be picked up by the vacuum gripper on the left arm (cf. \cref{fig:robot:left}).

It is worth noting how this specific task is particularly advantageous for the purposes of this work thanks to i) its simplicity and ii) the need for the human participant to perform the same actions multiple times.
We deliberately designed a task that does not require any particular skill from the builder, while being easy to understand and remember. Although it may be tedious for the user, the need of performing multiple actions of the same type is beneficial in terms of showcasing the online user adaptation capabilities introduced in \cref{sec:introduction}.
As detailed in \cref{sec:results} below, one of the assets of the proposed system is to be able to abide by the builder's preferences: in such a scenario, the robot is able to receive eventual negative feedback in case of wrong action, replan accordingly, and exhibit the effects of such replanning \textsl{within the same task execution}, i.e. without having to perform a new task from scratch.

\subsection{Experimental Evaluation} %
\label{sub:experimental_evaluation}

We demonstrate the proposed system in a live interaction with human participants. The robot is in charge of backing the user with the right supportive action at the right moment by virtue of a partial observation of the state of the world, the complete knowledge of the task execution plan, and the HTM-to-POMDP planner presented in \cref{sec:method}.
We devised two distinct experimental conditions, in a within-subjects design. For all the conditions, the skill set and capabilities of the robot do not vary, but the user preferences are explicitly altered, unbeknownst to the robot.
That is, the robot is exposed to a change in the state of the system---composed of the world plus the human---that it could not observe, but needs to deduce either by actively gathering information from the builder, or by building upon feedback coming from her.
In the following sections, the two experimental conditions are detailed. Please refer to \cref{sec:results} for a comparative evaluation.

\subsubsection{Condition A - full support} %
\label{ssub:exp_a_control_condition}

In this scenario, the robot expects to support the human to the best of its capabilities, that is by performing all the actions it is allowed to.
Firstly, the builder is introduced to the platform and the construction task.
The experimenter then proceeds to illustrate the Baxter's capabilities (i.e. providing parts, retrieving tools, holding objects and cleaning up the workspace) and the interaction channels the human is supposed to employ during task execution.
Next, the experimenter communicates to the user that the robot is supposed to perform all the supportive actions by itself, but also that the participant is free to take charge of some actions if she so chooses or if the robot fails.
No information is given in terms of what to expect from the robot, or how the human-robot interaction is supposed to occur.

\subsubsection{Condition B - adaptation to user preferences (no holding actions required)} %
\label{ssub:exp_b1_adaptation_condition_1}

This condition involves the same interaction between the human and the robot as Condition A.
As detailed in \cref{sub:experimental_evaluation}, the independent variable we tweak in our within-subjects experiments is the user preference for what concerns the support that the human participants expects from the robot.
In this scenario, the human worker is told to prefer not to have parts held by the robot while screwing.
Since the robot is unaware of this, it may still perform the holding action even if not required.
In case this happens, the human is instructed to negatively reward the robot by sending an error signal to the Baxter.

%
%

%

%
 %
%
\section{Results}
\label{sec:results}

The framework detailed in \cref{sec:method,sub:experimental_setup} has been released under the open-source LGPLv2.1 license, and is freely available on GitHub\footnote{ \href{https://github.com/scazlab/human_robot_collaboration}{\texttt{github.com/scazlab/human\_robot\_collaboration}} hosts the source code for the robot controllers, whereas \href{https://github.com/scazlab/task-models}{\texttt{github.com/ scazlab/task-models}} hosts the HTM to POMDP planner.}.
A number of C++ based ROS packages has been made available for robot-related software, whereas the planner has been encapsulated into a ROS-independent Python package.
In the following sections, we evaluate the proposed approach.
Firstly, we perform a series of off-line experiments to assess if the proposed model can derive effective policies against a variety of tasks and experimental conditions.
We show how our method outperforms ad-hoc policies on simulated interactions from these models (\cref{sub:evaluation_of_the_method}).
Lastly, we validate our system in a live interaction with human participants. We demonstrate how effective task policies can be computed that enable supportive behaviors during collaborative assembly tasks (\cref{sub:live_interaction}).

\subsection{Off-line evaluation of the method}
\label{sub:evaluation_of_the_method}

In this section, we present a quantitative evaluation of the proposed approach during off-line explorations.
To this end, we focus on the two most important aspects that are involved in the design of effective human-robot interactions: i) the flexibility of our method against a variety of task structures (\cref{ssub:task-structures}) and ii) its adaptability to custom user preferences (\cref{ssub:user_preferences}).

\subsubsection{Task structures}
\label{ssub:task-structures}

In order to demonstrate how the proposed method is able to provide effective support to a human partner in real-world collaborative tasks, we evaluate it on three task models, derived from distinct HTMs. The HTMs differ in the number of subtasks to solve, and in the type of relational operators between subtasks. They hence illustrate how we can derive policies from various task models.
All task models in this section are characterized by primitive subtasks that require a combination of: i) a set of tools that the robot needs to initially bring and then clean at the end of the task; ii) a shared supportive action `\textsl{a}'; iii) another supportive action, that can be either `\textsl{b}' or `\textsl{c}'. Let denote `B' the subtask that involves supportive actions `\textsl{a}' and `\textsl{b}' and `C' the one involving `\textsl{a}' and `\textsl{c}'.
The first HTM, denoted as \textsl{sequential} task, consists of a sequence of $20$  subtasks `B'.
The second task model is denoted as \textsl{uniform}; it consists of an alternative between $24$ subtasks, each composed of a sequence of four subtasks, each of type `B' or `C'. In other words, for each episode the current task is randomly chosen among any sequence of three subtask of type `B' or `C'.
The last HTM, denoted as \textsl{alternative} task, is an alternative between only $4$ sequences of four subtasks: `BCCC', `BBBB', `CBBC', and `CCBC'. It thus introduces a dependency between the successive required actions.

We compare the performance of the proposed approach against two hand-coded policies.
A \textsl{random} policy initially brings all the required tools (which is always a successful strategy); it then takes action `\textsl{a}', and after that it randomly chooses between action `\textsl{b}' and `\textsl{c}' until one succeeds. It finishes by cleaning the workspace.
When observing a failure (except on `\textsl{b}' and `\textsl{c}'), the policy simply repeats the last action.
The \textsl{repeat} policy is instead designed for the \textsl{sequential} task. Similarly to the \textsl{random} policy, it starts by providing the required tools, and then it repeats actions `\textsl{a}' and `\textsl{b}' $20$ times and then cleans the workspace. Similarly to \textsl{random}, it repeats failed actions until success.

\begin{figure}
  \centering
  \includegraphics[width=\linewidth]{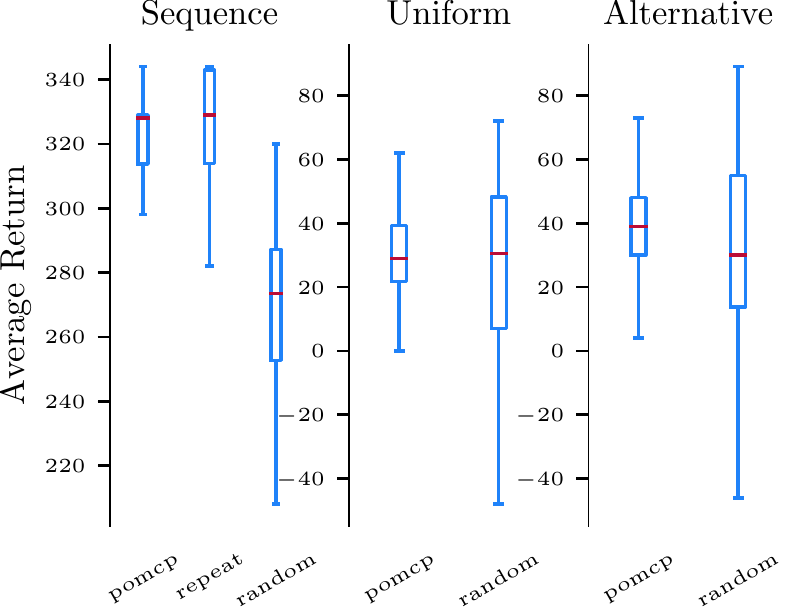}
  \caption{The \textsl{POMCP} policy matches or outperforms ad-hoc strategies against three different HTMs (\textsl{sequential}, \textsl{uniform}, and \textsl{alternative} task). The results are the distribution of returns over $100$ evaluation episodes from the two hand-coded policies (\textsl{repeat} and \textsl{random}) as well as the \textsl{POMCP} policy derived from the POMDP. On the last two tasks, the \textsl{repeat} policy fails in most of the cases and is stopped by an upper bound on the horizon of the episode. It gets a very low average return (between $-300$ and $-400$) that has been omitted from the figure to better compare the other policies.
  }
  \label{fig:result-multitask}
\end{figure}

\Cref{fig:result-multitask} presents the average return of each policy on the three conditions. Although \textsl{repeat} is very efficient on the \textsl{sequence} task, it is unfitted to the others and fails on the two other tasks. The \textsl{random} policy is suboptimal on all but the \textsl{uniform} task but can still solve them with a few failures.
On the other side, the \textsl{POMCP} policy that is learned from each task model matches or outperforms the others policies. The experiment hence demonstrates that we can leverage knowledge about the task structure to automatically derive efficient policies for each task.

\subsubsection{User Preferences} %
\label{ssub:user_preferences}

\begin{figure}
  \centering
  \includegraphics[width=\linewidth]{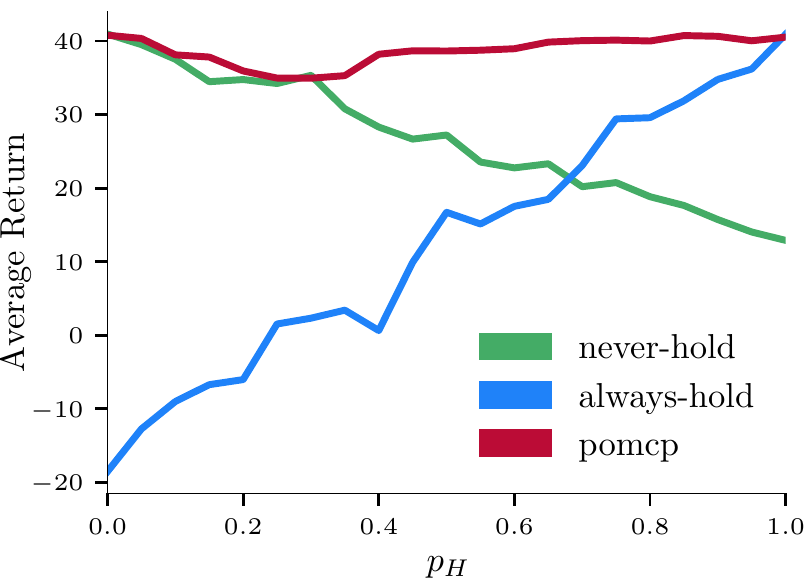}
  \caption{Comparison of the proposed strategy (red) against an `Always Hold' (green) and a `Never Hold` (blue) strategy during simulated interactions with varying degree of user preferences. For each of them, average return values and standard deviations with respect to the probability of the `Hold' preference $p_H$ are shown. The three strategies are tested against a single `Assemble Leg' subtask (see \cref{fig:htm:supportive}) with $20$ different values of $p_H$, ranging from $0.0$ (i.e. the human never wants the robot to hold) to $1.0$ (i.e. the human always like the robot to hold). For each of the $20$ different preferences, the results are averaged from $100$ simulated interactions, for a total of $6000$.}
  \label{fig:result-preferences}
\end{figure}

Being able to comply with---and adapt to---custom user preferences is crucial for a robot that needs to provide the best support to its partner.
A prompt and personalized response allows for a more natural interaction and a less cognitively demanding execution, which ultimately result in a more efficient collaboration.
To this end, we present a system that is successfully able to account for user preferences. As detailed in \cref{sub:experimental_evaluation}, the participant is allowed to choose if the robot should provide support by holding during screwing or not.
We compare our approach against two hard-coded policies: the most proactive strategy (i.e. always offering to hold for support) and the most conservative one (i.e. never proposing to support the human). \Cref{fig:result-preferences} demonstrates that these strategies are only optimal when they match with the expected user preference; in the intermediate scenarios, their performance degrades quickly as the uncertainty on the actual user preference increases.
Conversely, our system is able to adapt to whether the human partner would like the robot to provide hold support or not, and outperforms both strategies in the majority of conditions.

\subsection{Live interaction with human participants}
\label{sub:live_interaction}

\begin{figure*}
\centering
  \subfloat[Condition A]{\includegraphics[width=.315\textwidth]{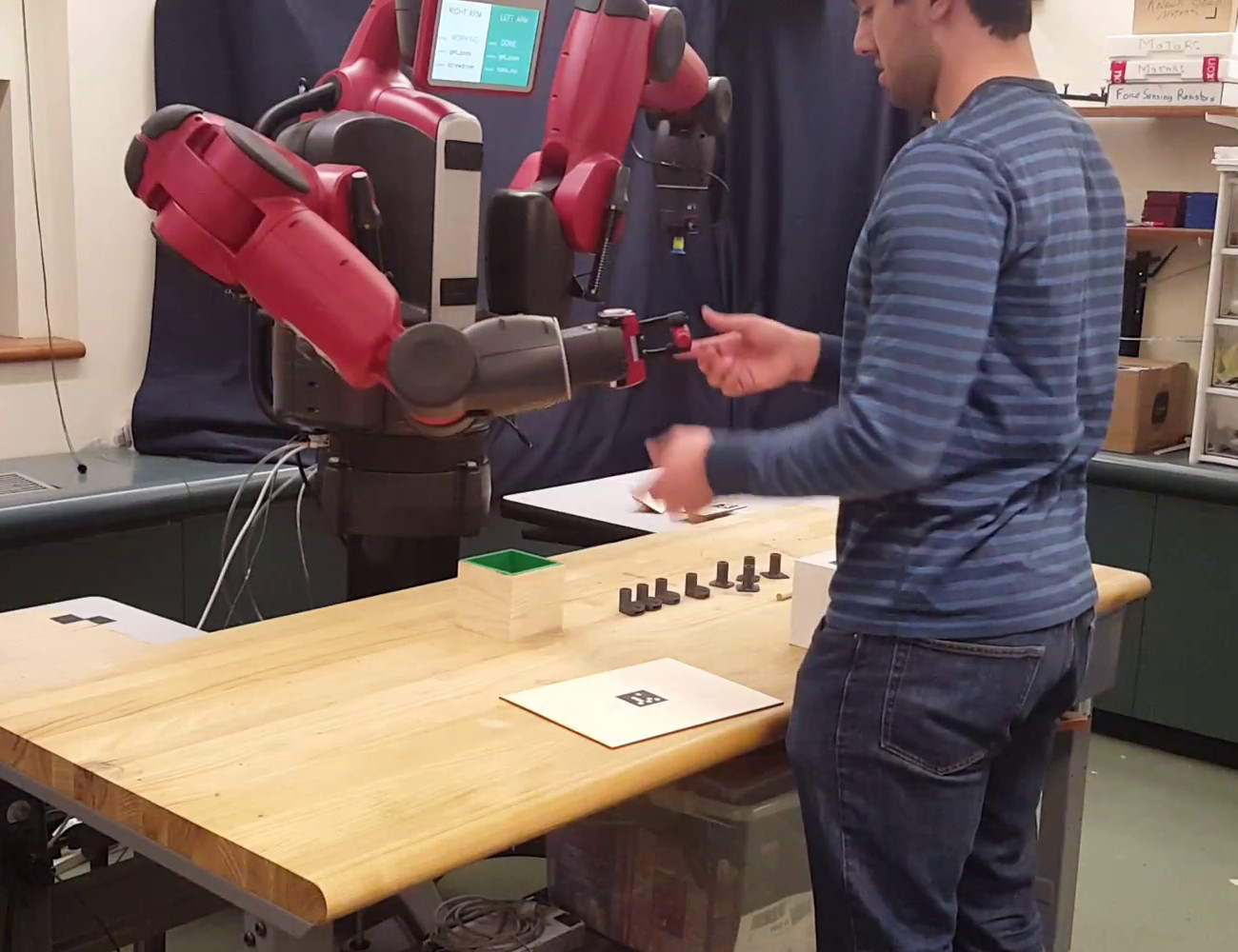}\quad }
  \subfloat[Condition A]{\includegraphics[width=.315\textwidth]{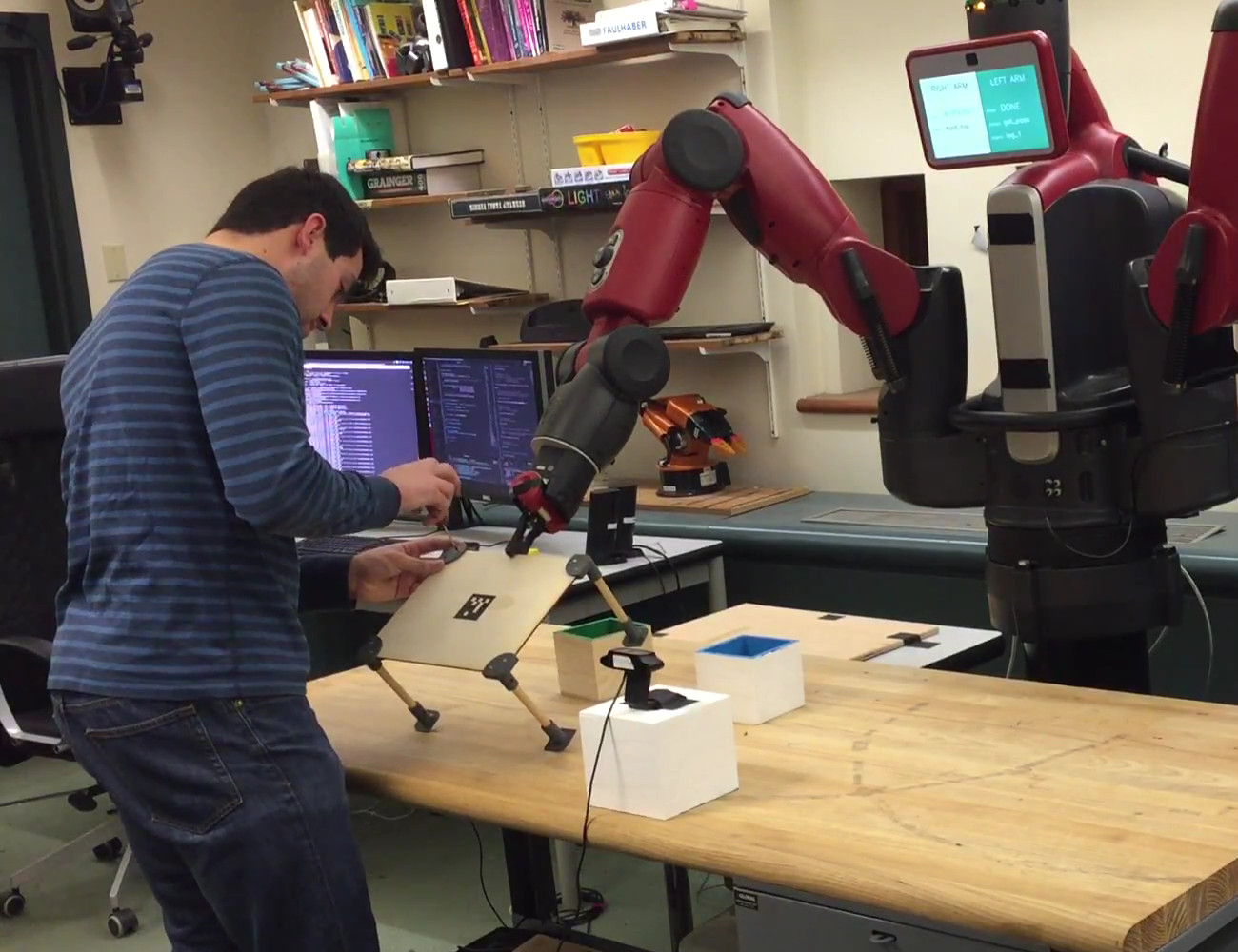}\quad }
  \subfloat[Condition A]{\includegraphics[width=.315\textwidth]{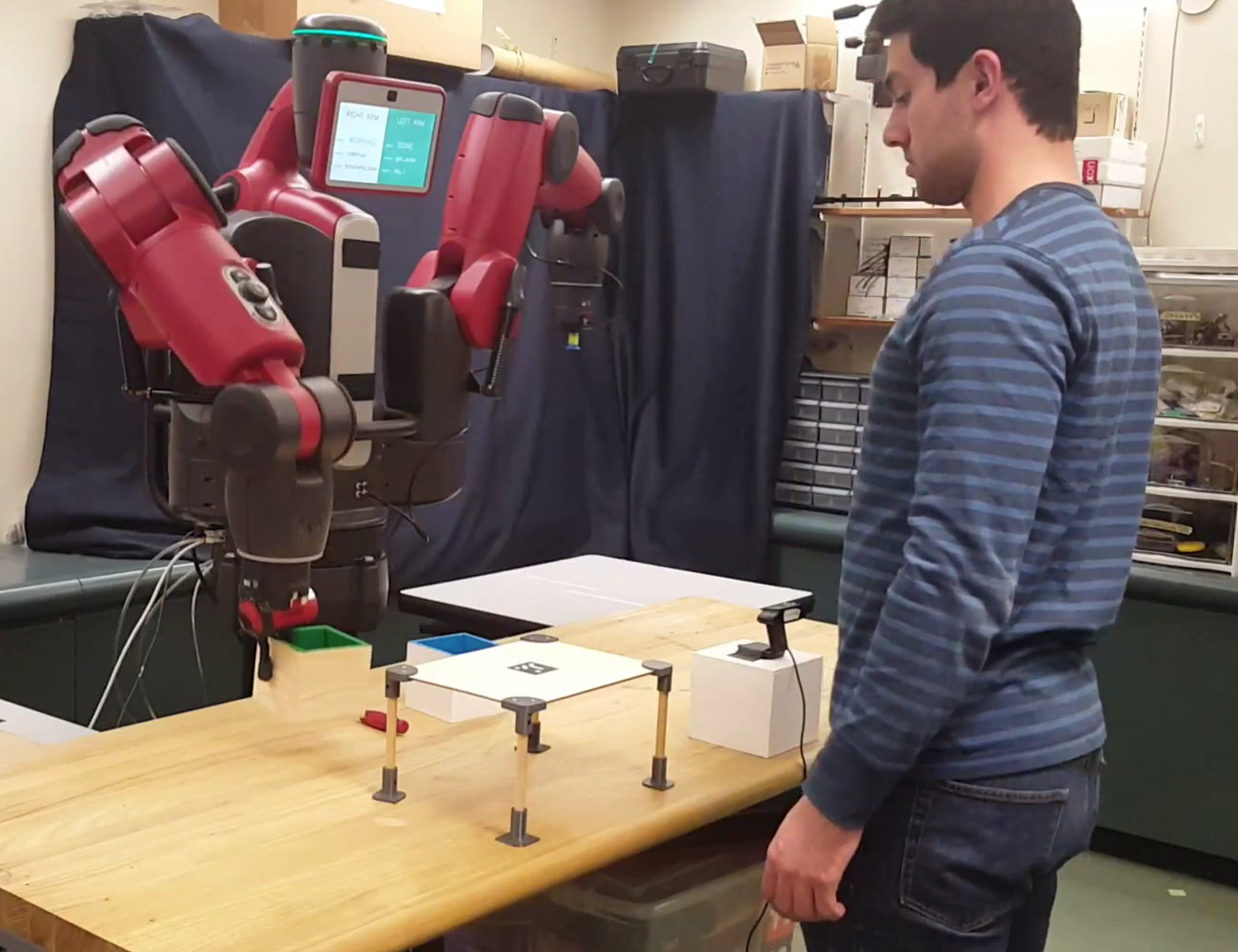}}\\
  \subfloat[Condition B]{\includegraphics[width=.315\textwidth]{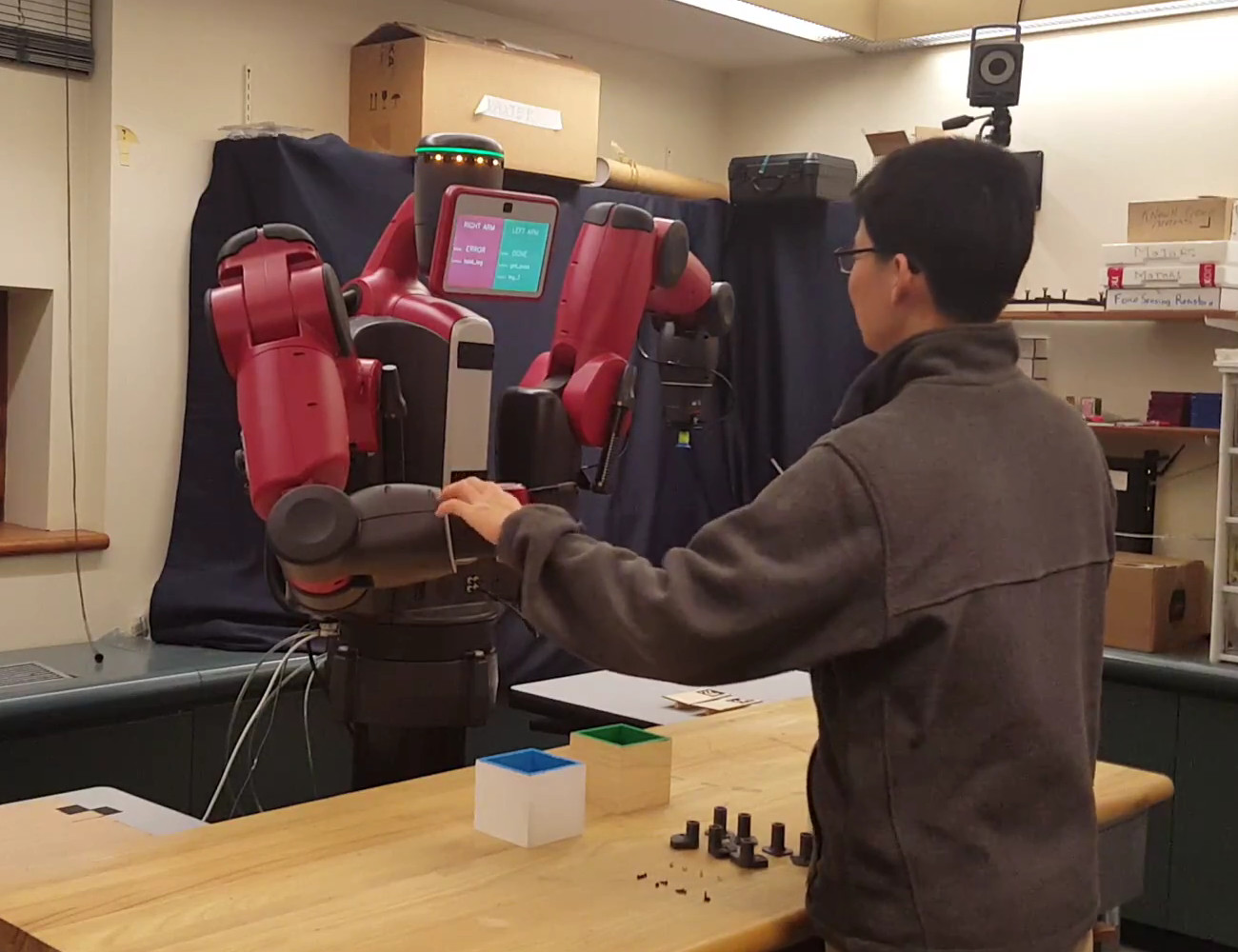}\quad \label{fig:stills:b1}}
  \subfloat[Condition B]{\includegraphics[width=.315\textwidth]{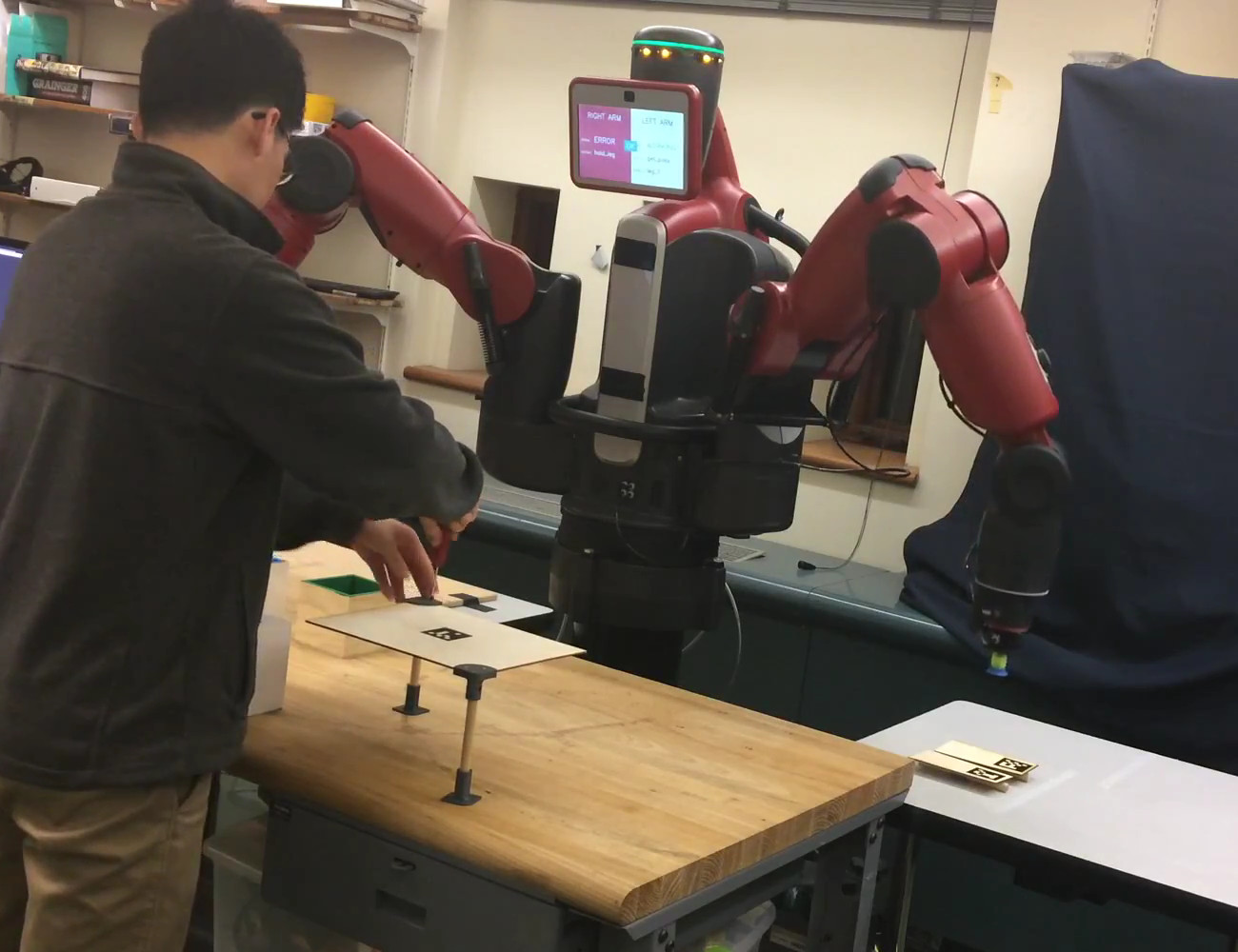}\quad }
  \subfloat[Condition B]{\includegraphics[width=.315\textwidth]{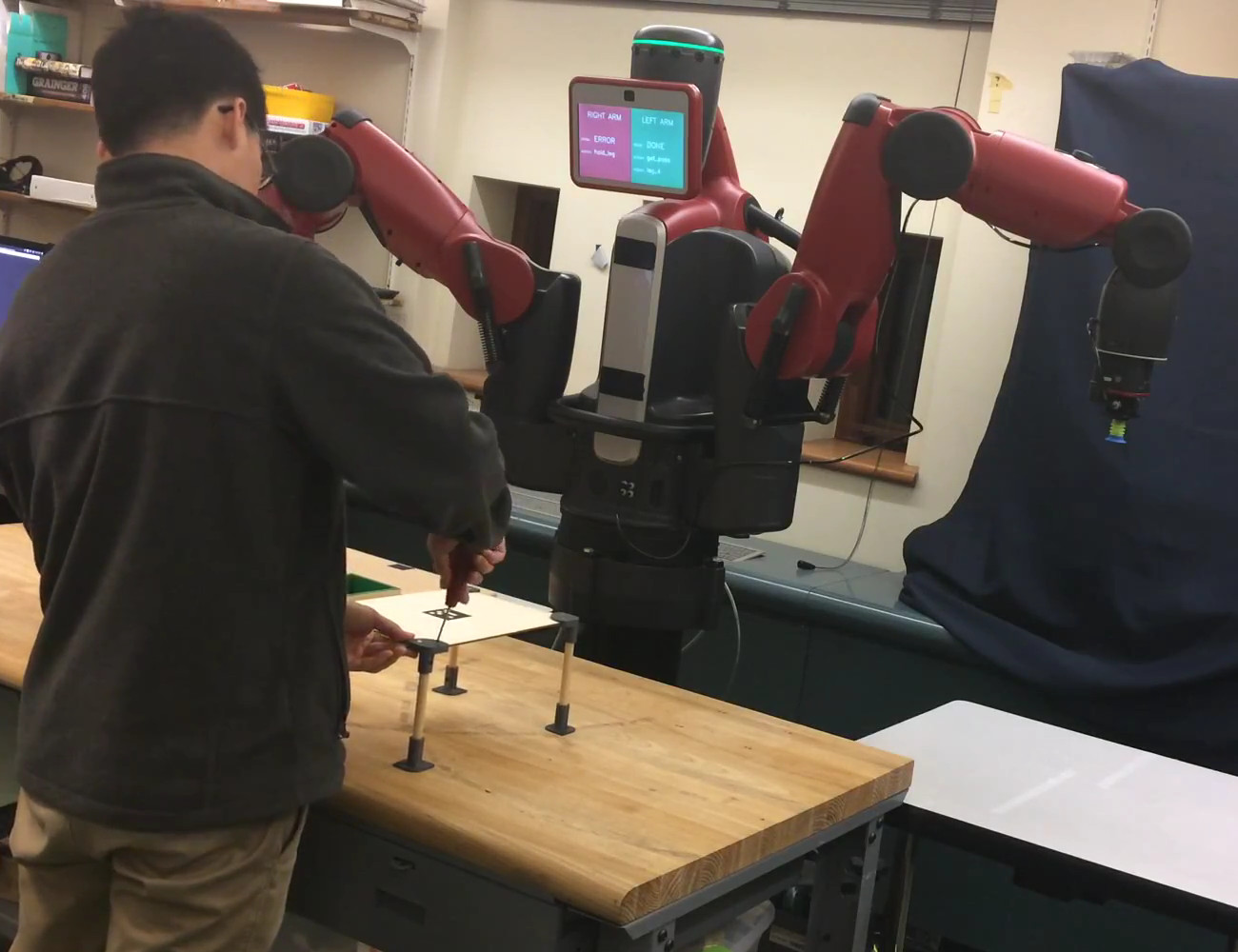}}\\
  \caption{Snapshots acquired during the collaborative assembly of the table in Condition A (full support, top) and Condition B (no holding required, bottom). Condition A (top):
  \textbf{a)} Baxter provides the tool to the participant;
  \textbf{b)} the robot supports the human by holding the tabletop while the human screws the leg in place;
  \textbf{c)} the user has finished his task and observes the robot freeing the workspace from the box of linkages.
  Condition B (bottom):
  \textbf{a)} The user signals to the robot that the holding action is not required by pressing the error button, and the robot acknowledges back by signaling into its display that it received this information;
  \textbf{b)} The hold action is not performed any more, but other actions such as the retrieval of the leg are still performed;
  \textbf{c)} The human participant completes the execution of the task without the help of the robot, as required.}  \label{fig:stills}
\end{figure*}

We demonstrate the proposed framework in an experimental scenario where a human participant is engaged in the joint construction of toy furniture with the Baxter robot.
We define the problem detailed in \cref{sub:experiment_design} as a sequence of eight subtasks, whose HTM is shown in \cref{fig:htm:supportive}.
All the subtasks depicted in figure require high dexterity and perception skills, and thus need to be performed by the human builder exclusively. For each of the four legs, the builder is in charge of firstly screwing the linkages (bracket and foot) onto the leg, and subsequently screwing the bracket onto the tabletop.
As introduced in \cref{sub:experiment_design}, retrieving parts and tools from their respective pool has been designed as one of the supportive actions the helper robot can choose from. Further, the Baxter robot is allowed to hold parts in order to facilitate the participant's work, and to clean up the workstation when it deems it appropriate.
For more information about the actual interaction, we refer the reader to the accompanying video, which has been summarized in \cref{fig:stills} (full resolution available at \href{https://youtu.be/OEH-DvNS0e4}{\texttt{youtu.be/OEH-DvNS0e4}}).

{%
\begin{center}
  \begin{table}
    \caption{Evaluation of the human-robot collaboration. For both conditions, average completion time and average number of robot actions are shown. See text for details.}
    \footnotesize \centering
    \begin{tabular}{R{2cm}cc}
      \toprule
       & \textbf{Condition A -- \textsl{hold}} & \textbf{Condition B -- \textsl{no-hold}} \T \B \\
      \cmidrule(r){2-2} \cmidrule(r){3-3} \T \B
      Average Completion Time        & $649$ [s] & $747$ [s] \T \B \vspace{4pt} \\
      Avg. Number of Robot Actions   &  $21$     &  $14$     \T \B \vspace{4pt} \\
      \bottomrule
    \end{tabular}
    \label{tab:results}
  \end{table}
\end{center}
}

The system has been evaluated with four participants; each participant performed the task in both Condition A and B, in a within-subject design (please refer \cref{sub:experimental_evaluation} for a description of the experimental conditions).
In all the demonstrations performed, the robot was successful in providing support to the human.
As shown in \cref{tab:results}, Condition A, when the robot was allowed to provide more support to the human---and was thus intervening more in the task, we register an overall reduction of task completion time ($13.2\%$ on average).

{%
\begin{center}
  \begin{table}
    \caption{Example histories of actions and observations during the interaction, for the two conditions. $\hat{p_H}$ is the estimation of the probability for the `hold' preference in the robot's internal belief.}
    \footnotesize \centering
    \begin{tabular}{cccccc}
      \toprule
      \multicolumn{3}{c}{\textbf{Condition A -- \textsl{hold}}} & \multicolumn{3}{c}{\textbf{Condition B -- \textsl{no-hold}}} \T \B \\
      actions & obs & $\hat{p_H}$ & actions & obs & $\hat{p_H}$ \T \B \\
      \cmidrule(r){1-3} \cmidrule(l){4-6} \T \B
      bring screws               &          none & $.34$ & bring screws      &          none & $.43$ \T \B \\
      bring leg                  &          none & $.38$ & bring leg         &          none & $.46$ \T \B \\
      \color{NavyBlue}\textbf{bring screwdriver} & \color{NavyBlue}\textbf{fail} & \color{NavyBlue}$.36$ & bring screwdriver &          none & $.44$ \T \B \\
      bring top                  &          none & $.37$ & bring joints      &          fail & $.38$ \T \B \\
      bring joints               &          none & $.39$ & bring top         &          none & $.38$ \T \B \\
      \color{NavyBlue}\textbf{bring screwdriver} & \color{NavyBlue}\textbf{fail} & \color{NavyBlue}$.43$ & bring joints      &          none & $.35$ \T \B \\
      \color{NavyBlue}\textbf{bring screwdriver} & \color{NavyBlue}\textbf{none} & \color{NavyBlue}$.42$ & \color{BrickRed}\textbf{hold}     & \color{BrickRed}\textbf{fail} & \color{BrickRed}$.03$ \T \B \\
      \color{Green}\textbf{hold}              & \color{Green}\textbf{none} & \color{Green}$.41$ & bring leg         &          none & $.03$ \T \B \\
      \color{Green}\textbf{hold}              & \color{Green}\textbf{none} & \color{Green}$.86$ & \color{BrickRed}\textbf{wait}     & \color{BrickRed}\textbf{none} & \color{BrickRed}$.02$ \T \B \\
      bring leg                  &          none & $.97$ & \color{BrickRed}\textbf{wait}     & \color{BrickRed}\textbf{none} & \color{BrickRed}$.01$ \T \B \\
      \color{Green}\textbf{hold}              & \color{Green}\textbf{none} & \color{Green}$.97$ & wait              &          none & $.01$ \T \B \\
      hold                       &          none & $.99$ & bring leg         &          none & $.01$ \T \B \\
      bring leg                  &          none & $1.0$ & wait              &          none & $.01$ \T \B \\
      hold                       &          none & $1.0$ & wait              &          none & $.02$ \T \B \\
      hold                       &          none & $1.0$ & bring leg         &          none & $.03$ \T \B \\
      bring leg                  &          none & $1.0$ & wait              &          none & $.03$ \T \B \\
      hold                       &          none & $1.0$ & wait              &          none & $.04$ \T \B \\
      hold                       &          none & $1.0$ & clear joints      &          none & $.04$ \T \B \\
      clear screws               &          none & $1.0$ & clear screws      &          none & $.04$ \T \B \\
      clear joints               &          none & $1.0$ & clear screwdriver &          none & $.01$ \T \B \\
      clear screwdriver          &          none & $1.0$ & wait              &          none & $.01$ \T \B \\
      wait                       &          none & $1.0$ &                   &      &       \T \B \\
       \bottomrule
    \end{tabular}
    \label{tab:histories}
  \end{table}
\end{center}
}

More interesting is to evaluate the extent to which the planner is able to recover from robot failures and policy errors.
To this end, we highlight two example trajectories in \cref{tab:histories}.
They correspond to actual trajectories, one from Condition A, where the robot is expected to hold the parts, and one from Condition B, where the user will signal her preference about not wanting parts to be held.
The interaction channels described in \cref{sub:experimental_setup} allow for a certain degree of flexibility for what concerns the type of communication the builder and the helper can engage into.
In particular, the Baxter is allowed to gather information about the user's preference by taking a hold action although unsure about it. This would ultimately disambiguate its uncertainty, because the builder would communicate failure in case her preference is a `no-hold'.
For the purposes of this work, we examine here three prototypical interactions, color-highlighted in~\cref{tab:histories} with their corresponding preference update in the third column of each table:

\begin{itemize}
\item[--] \textsl{Robot-initiated failure}: as described in \cref{sec:method}, the robot is not allowed to directly perceive the state of the world and the progression of the task. Still, it is possible for it to detect an action failure by using its own internal sensors.
If this is the case, e.g. when the robot tries to pick up the screwdriver but the gripper is empty, the system is able to re-plan its execution and re-schedule the action at a later stage. As highlighted  in \cref{tab:histories}, blue sequence, the robot does not necessarily repeat the same action right after the failure is detected, since other actions may have similar priority at that point.
\item[--] \textsl{Successful hold action} (green sequence in \cref{tab:histories}): the robot starts with a non-zero estimation $\hat{p_H}$ of probability for the hold preference. If, while performing an `hold' action, it does not receive negative feedback from the user (observation: `none'), $\hat{p_H}$ increases as it becomes more likely that the builder wants the robot to hold. This is further enforced with subsequent `hold' actions.
\item[--] \textsl{User-initiated failure}: in Condition B, the system experiences an user-initiated failure while proposing to hold the part (observation: `fail', red sequence in \cref{tab:histories}). As a consequence of this, the probability of the hold action in the belief distribution $\hat{p_H}$ decreases, and the robot will not perform this action in the future.
Rather than hold parts for the user, it will wait for her to complete the action, and will move on to the next step when she communicates completion of the subtask.
\end{itemize}

One last aspect worth elaborating on is the fact that, in order for the robot to be perceived as an effective collaborator, transparency of the system during interaction is paramount. The human needs to be able to access (to a certain degree) what the robot's internal state is, what it thinks about the task progression and, importantly, how it intends to act next. Failure to deliver transparency results in user frustration and task inefficiency.
Within this context, the overlapping, redundant interaction channels (cf. \cref{sub:experimental_setup}) were beneficial in guaranteeing a transparent exchange of information between the two partners.
This is particularly important in case of unexpected deviations from the robot's nominal course of actions---that is, robot failures. Both in case of a robot-initiated failure and a human-initiated error signal (\cref{fig:stills:b1}), the system was able to acknowledge the user about its error state through the Baxter's head display and/or speech utterances.
In this way, it was always evident to the user that the robot failed, and eventually why it failed (in case the failure was not of robot-initiated).

\section{Discussion and Future Work} %
\label{sec:discussion}

In this work, we present a system able to convert high-level hierarchical task representations into low-level robot policies. We demonstrate robustness to task representations with varying complexity, as well as a certain degree of customization with respect to task-relevant variables such as user preferences and task completion time.
Further, we introduce a novel experimental design, composed of flexible and modular constituent parts that can be easily reconfigured for a variety of different experimental scenarios.
Finally, we provide demonstration of our technique in a mixed-initiative human-robot collaboration.
As evident in the accompanying video, the human maintains full control throughout the task execution, but the robot acts independently, anticipates human needs, and does not wait to be told what to do.
As mentioned in \cref{sec:introduction}, the paradigm in which we operate is neither to attempt implementing the ideal system that never fails and does not contemplate the occurrence of failures, nor shaping the environment in such a way that it prevents the robot to fail. Rather, we decidedly embrace the idea that robots' perception and actions are inherently faulty, and errors during operation are possible and \textsl{expected to occur}.
The approach we present does not intend to compete with the optimized assembly lines that takes months to design, but provides an easy-to-deploy, reconfigurable paradigm, suitable for small and medium enterprises.

To our knowledge, this work is the first attempt at a \textsl{practical} demonstration of supportive behaviors in a realistic human-robot collaborative scenario.
In addition, we fundamentally differ from past research on the topic, where collaboration typically translates to the human and the robot tasked with parallel, non overlapping subtasks and rigid, structured interactions.
Rather, our experiment shows a fully integrated interaction, where the human and the robot \textsl{physically engage in shared-environment collaboration}.

A more extensive evaluation of the scalability of the proposed framework to a broader domain of applications is the major direction for future work.
Although the simulated interactions shown in \cref{ssub:task-structures} proved its feasibility in theory, it remains to be seen how much the approach can scale up to more complex tasks in practice.
In particular, we plan on leveraging the flexibility of the HTM representation to: i) model more complex task structures; ii) apply the method to different interaction domains.
Furthermore, our previous work introduced a model that allows the robot to effectively exploit basic communication capabilities in order to target the problem of task allocation and information gathering~\cite{Roncone2017}. Bringing this level of interaction in the current setup is also a direction of convergence.

In this work, we assume that the high level representations our system relies on are either already available or easy enough to generate. It is however a matter of future work to explore how this assumption holds in various application domains, and whether sufficiently precise models can be learned from spoken instructions or user demonstrations.
Our method also relies on existing controllers for the supportive actions. Interesting directions of future work would entail the ability of users to teach new primitives to the robot (e.g. by demonstration) and then combine these primitives into more complex task models.

One of the limitations of the live interaction experiment shown in \cref{sub:live_interaction} is that its evaluation is qualitative in nature.
Nonetheless, the proposed HTM-to-POMDP method is sound, and previous work demonstrated how the same technique applied to a different PODMP allows for statistically significant results in terms of overall completion time~\cite{Roncone2017}.
More interestingly, the system shown here opens the door to a wide array of user studies to assess the quality and the effectiveness of the interaction between the human and the robot.
The degree of proaction shown by the robot during collaboration can significantly lower the barrier to entry for non-expert users, in that users can immediately see what the robot is capable of, without over- or under-estimating its skill set.
In this regard, an extensive user study would help solidifying this intuition, and assessing how useful the proposed system is in setting expectations for naive users.
Finally, although our prior work showed a general user preference toward our system~\cite{Roncone2017}, a broader user study would prove statistical significance in terms of reduced levels of stress and cognitive load to the user. Importantly, this would also allow to highlight potential friction points that can be leveraged to better design the interaction.

\bibliographystyle{plainnat}
\bibliography{tro_17}

\begin{IEEEbiography}[{\includegraphics[width=1in,clip,keepaspectratio]{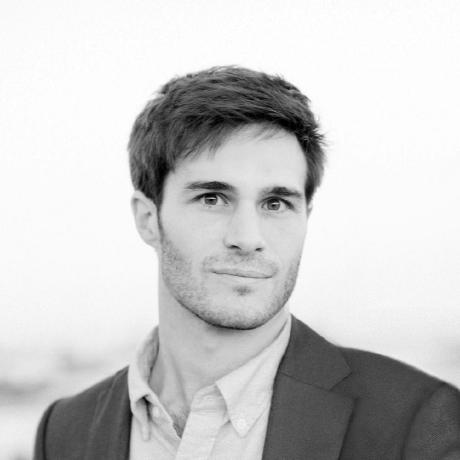}}]{Olivier Mangin}
is a developmental and social robotics researcher, now a postdoctoral associate in the Social Robotics laboratory at Yale university. His interest lies in understanding how structure may appear in the sensorimotor interaction between an infant or robot and its environment. In particular he studies the development of collaborative and linguistic behaviors from observation, imitation or self exploration.
Olivier completed his PhD at INRIA in Bordeaux, under the supervision of Pierre-Yves Oudeyer, after a master degree in machine learning.
 \end{IEEEbiography}

\begin{IEEEbiography}[{\includegraphics[width=1in,clip,keepaspectratio]{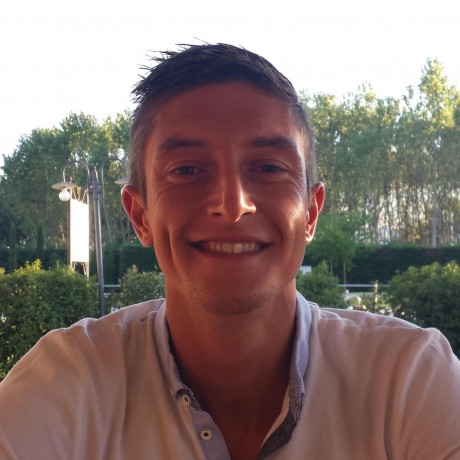}}]{Alessandro Roncone}
is a Postdoctoral Associate at the \href{http://scazlab.yale.edu/}{Social Robotics Lab} in Yale University. The central motivating theme of his research is endowing robots with enough communication, perception and control capabilities to enable close, natural and extended cooperation with humans.
He completed his Ph.D. in Robotics, Cognition and Interaction Technologies from the \href{https://iit.it/}{Italian Institute of Technology}, working on a bio-inspired peripersonal space model that improved the sensorimotor capabilities of the \href{http://www.icub.org/}{iCub} humanoid robot.
His current research focuses on the exploitation of bidirectional communication between the robot and the human in the context of human-robot collaboration and advanced manufacturing.
 \end{IEEEbiography}

\begin{IEEEbiography}[{\includegraphics[width=1in,clip,keepaspectratio]{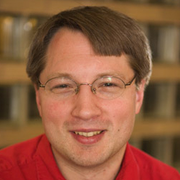}}]{Brian Scassellati}
is a Professor of Computer Science, Cognitive Science, and Mechanical Engineering at Yale University and Director of the NSF Expedition on Socially Assistive Robotics. His research focuses on building embodied computational models of human social behavior, especially the developmental progression of early social skills. Using computational modeling and socially interactive robots, his research evaluates models of how infants acquire social skills and assists in the diagnosis and quantification of disorders of social development (such as autism). His other interests include humanoid robots, human-robot interaction, artificial intelligence, machine perception, and social learning.
 \end{IEEEbiography}

\end{document}